\documentclass{article}

    \usepackage[preprint,nonatbib]{neurips_2021}

\usepackage{color}
\usepackage[dvipsnames]{xcolor}
\usepackage{epsfig}
\usepackage{graphicx}

\usepackage[export]{adjustbox}
\usepackage{arydshln}
\usepackage{colortbl}
\usepackage{float,wrapfig}
\usepackage{hhline}
\usepackage{multirow}
\usepackage{tabularx}

\usepackage{amsmath,amsfonts,amsthm,amssymb}
\usepackage{bm}
\usepackage{nicefrac}
\usepackage{microtype}
\usepackage{xfrac}

\usepackage{changepage}
\usepackage{extramarks}
\usepackage{fancyhdr}
\usepackage{lastpage}
\usepackage{setspace}
\usepackage{soul}
\usepackage{xspace}

\usepackage[pagebackref=true,breaklinks=true,colorlinks]{hyperref}
\usepackage[nocompress]{cite}
\usepackage{url}

\usepackage{enumerate}

\usepackage{etoolbox,siunitx}
\usepackage{afterpage}
\usepackage{subcaption}
\usepackage[title]{appendix}

\usepackage{extarrows}
\usepackage{ragged2e}

\newcommand{\figref}[1]{\figurename~\ref{#1}}

\newcommand{\subsecref}[1]{Subsection~\ref{#1}}
\newcommand{\eqnref}[1]{Eqn.~\eqref{#1}}
\newcommand{\tblref}[1]{Table~\ref{#1}}

\newcommand{\ignorethis}[1]{}

\makeatletter
\DeclareRobustCommand\onedot{\futurelet\@let@token\@onedot}
\def\@onedot{\ifx\@let@token.\else.\null\fi\xspace}

\def\eg{\emph{e.g}\onedot} 
\def\ie{\emph{i.e}\onedot}

\def\etal{\emph{et al}\onedot}
\makeatother

\newcommand*{\rom}[1]{\expandafter\romannumeral #1}

\definecolor{mydarkblue}{rgb}{0,0.08,1}
\definecolor{mydarkgreen}{rgb}{0.02,0.6,0.02}
\definecolor{mydarkred}{rgb}{0.8,0.02,0.02}
\definecolor{mydarkorange}{rgb}{0.40,0.2,0.02}
\definecolor{mypurple}{RGB}{111,0,255}
\definecolor{myred}{rgb}{1.0,0.0,0.0}
\definecolor{mygold}{rgb}{0.75,0.6,0.12}
\definecolor{myblue}{rgb}{0,0.2,0.8}
\definecolor{mydarkgray}{rgb}{0.66,0.66,0.66}

\newif\ifcolor
\colortrue

\newcommand{\nuke}[1]{\textcolor{black}{#1}}
\newcommand{\workwork}[1]{\textcolor{black}{#1}}

\newcommand{\dcc}[1]{}
\newcommand{\yn}[1]{\textcolor{black}{#1}}
\newcommand{\ync}[1]{}
\newcommand{\rg}[1]{\textcolor{black}{#1}}
\newcommand{\rgc}[1]{}

\newcommand{\w}{$\mathcal{W}$\xspace}
\newcommand{\wplus}{$\mathcal{W}+$\xspace}
\newcommand{\s}{$\mathcal{S}$\xspace}

\newcommand{\ccbync}{\href{https://creativecommons.org/licenses/by-nc/4.0/legalcode}{CC BY-NC 4.0}}

\newcommand{\cczero}{\href{https://creativecommons.org/publicdomain/zero/1.0/}{CC0 1.0}}

\newcommand{\ccbyncsa}{\href{https://creativecommons.org/licenses/by-nc-sa/4.0/}{CC BY-NC-SA 4.0}}

\newcommand{\nvsrc}{\href{https://nvlabs.github.io/stylegan2/license.html
}{Nvidia Source Code License-NC}}

\newcommand{\bsd}{\href{https://opensource.org/licenses/BSD-3-Clause}{BSD 3-Clause}}

\newcommand{\mitlic}{\href{https://opensource.org/licenses/MIT}{MIT License}}

\DeclareGraphicsExtensions{.jpg, .pdf, .png, .gif}

\title{LARGE: Latent-Based Regression through GAN Semantics}

\makeatletter
\newcommand{\printfnsymbol}[1]{%
  \textsuperscript{\@fnsymbol{#1}}%
}
\makeatother

\author{%
  Yotam Nitzan\thanks{Indicates equal contribution} \\
  Tel-Aviv University \\
   \And
   Rinon Gal\printfnsymbol{1} \\
   Tel-Aviv University \\
   \And
   Ofir Brenner \\
   Tel-Aviv University \\
   \And
   Daniel Cohen-Or \\
   Tel-Aviv University
   \AND
   \href{https://yotamnitzan.github.io/LARGE}{yotamnitzan.github.io/LARGE}
   \vspace{-2em}
}

\begin{document}

\maketitle

\begin{abstract}
   We propose a novel method for solving regression tasks using few-shot or weak supervision. At the core of our method is the fundamental observation that GANs are incredibly successful at encoding semantic information within their latent space, even in a completely unsupervised setting. For modern generative frameworks, this semantic encoding manifests as smooth, linear directions which affect image attributes in a disentangled manner. These directions have been widely used in GAN-based image editing.
   We show that such directions are not only linear, but that the magnitude of change induced on the respective attribute is approximately linear with respect to the distance traveled along them. By leveraging this observation, our method turns a pre-trained GAN into a regression model, using as few as two labeled samples. This enables solving regression tasks on datasets and attributes which are difficult to produce quality supervision for. Additionally, we show that the same latent-distances can be used to sort collections of images by the strength of given attributes, even in the absence of explicit supervision. 
  Extensive experimental evaluations demonstrate that our method can be applied across a wide range of domains, leverage multiple latent direction discovery frameworks, and achieve state-of-the-art results in few-shot and low-supervision settings, even when compared to methods designed to tackle a single task.
   
\end{abstract}
\section{Introduction}

In recent years, Generative Adversarial Networks (GANs) \cite{goodfellow2014generative} have been at the forefront of deep learning research. GANs revolutionized countless generative tasks, such as unconditional image synthesis \cite{karras2020analyzing, brock2018large}, cross-domain image-to-image translation \cite{isola2017image,zhu2017unpaired} and super-resolution \cite{ledig2017photo}. On many such tasks, GAN-based methods have achieved unprecedented success in terms of visual quality, fidelity and diversity.
Beyond generative tasks, numerous works have proposed to use GANs for downstream discriminative objectives, such as classification. Their shared premise is that a generator can synthesize novel samples - often in a controllable manner. These generated samples can then serve as a dataset for training models for downstream tasks.
While this approach appears promising on paper \cite{tanaka2019data, salimans2016improved, antoniou2017data, mariani2018bagan}, this simple idea has enjoyed limited success \cite{cubuk2020randaugment, ravuri2019classification}.

We propose an alternative approach to harnessing the rapid advancement of GANs for downstream tasks. Specifically, we deviate from previous attempts to generate or augment training data. Instead, we focus on extracting information from the incredibly well-behaved latent space of modern GAN architectures, and specifically StyleGAN \cite{karras2019style, karras2020analyzing}. The latent spaces of StyleGAN have been studied extensively \cite{xia2021gan}, and were shown to be highly semantic, a property which led to their wide use across a range of generative tasks. 

\begin{figure}
\setlength{\tabcolsep}{1pt}
    \centering
    \begin{tabular}{c c}
        \includegraphics[width=0.45\linewidth]{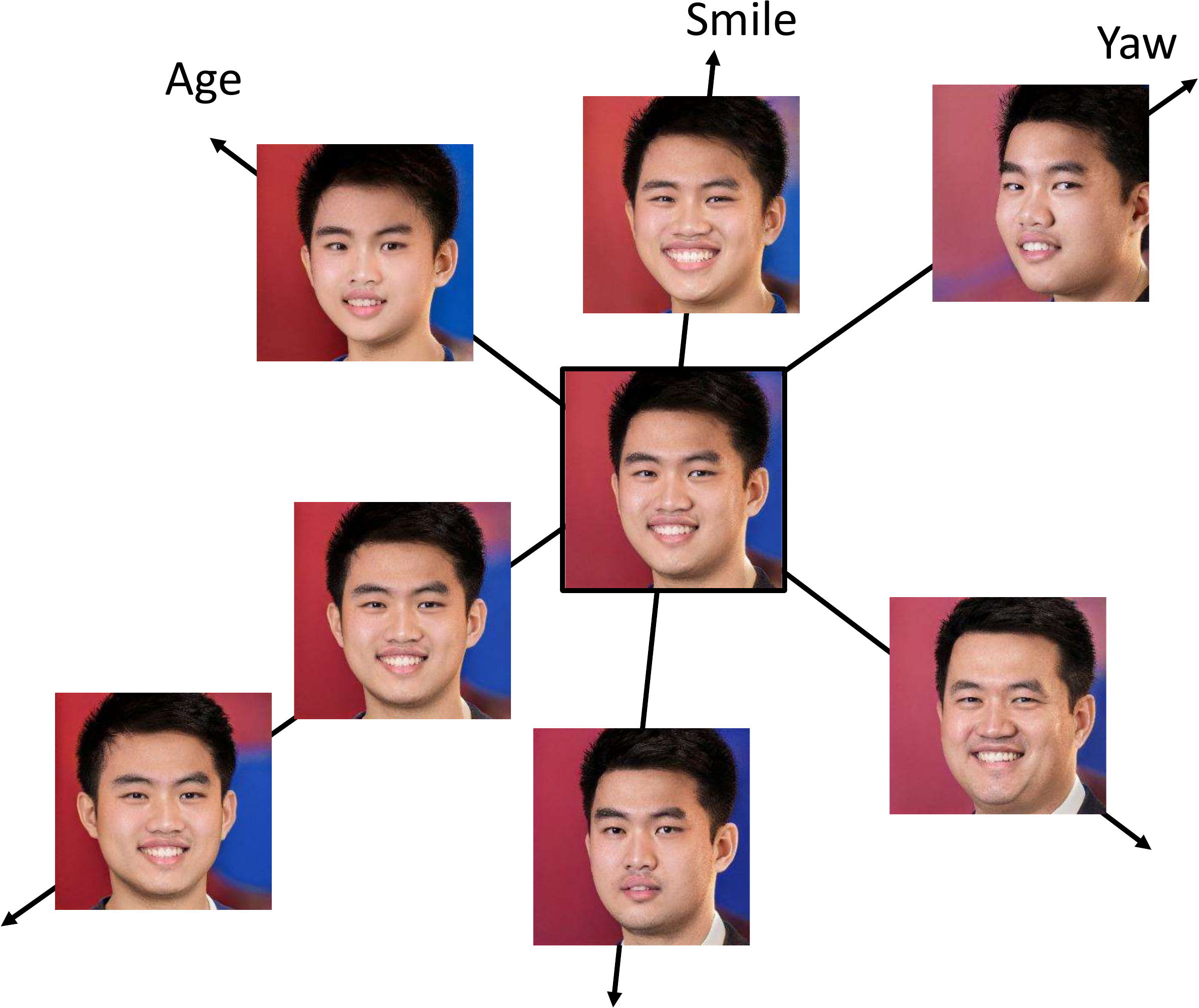} &
        \includegraphics[width=0.55\linewidth]{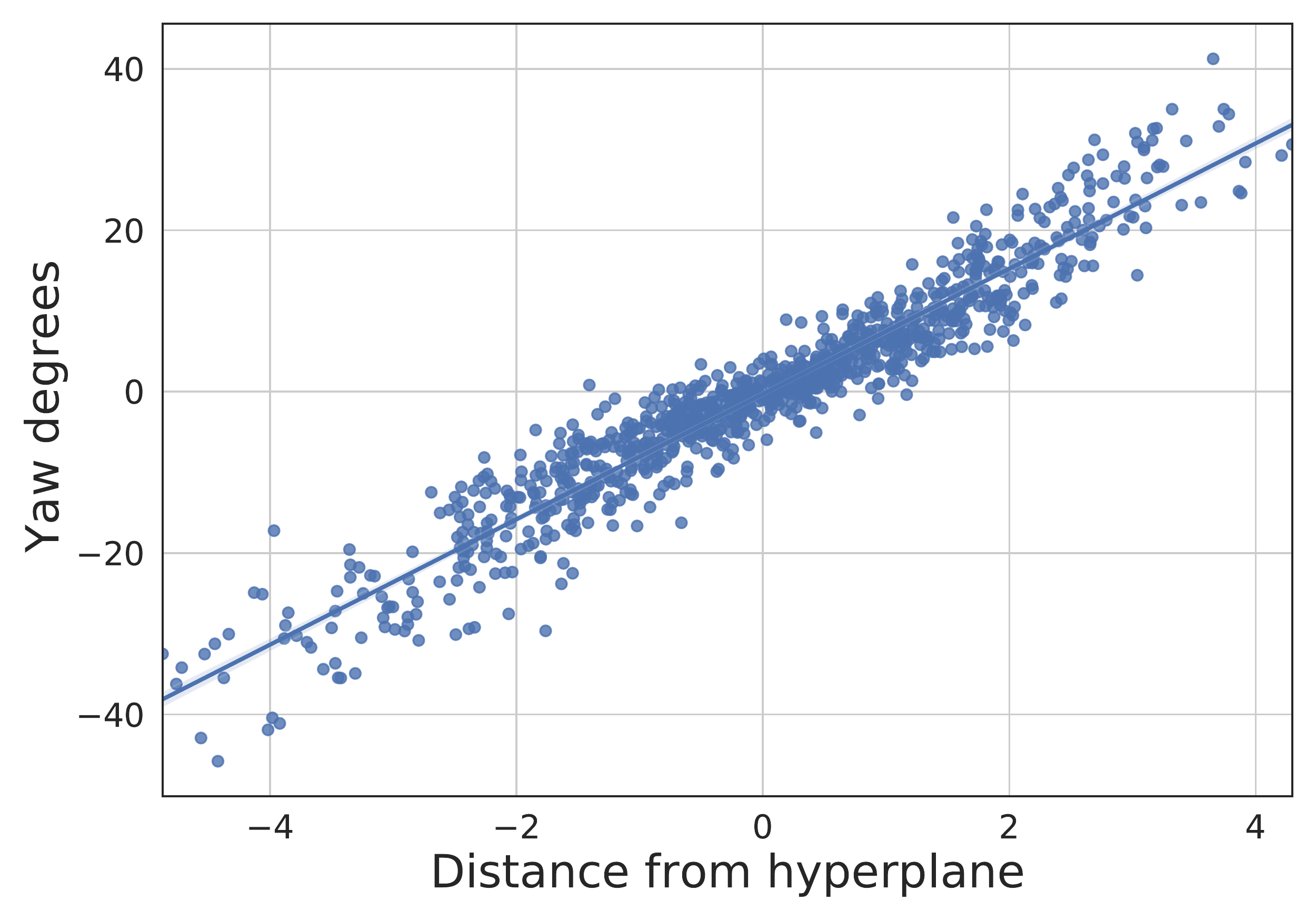} \\
        (a) & (b)
    \end{tabular}
    \vspace{-0.2cm}
    \caption{The two linearities in a GAN's latent space. (a) A 2D illustration depicting the \textbf{linear} latent directions corresponding to different semantic properties. 
    (b) \rg{Yaw angles \cite{zhou2020whenet} of generated face images as a function of the distance of their corresponding latent codes from a yaw hyperplane \cite{shen2020interpreting}.} There exists an approximately \textbf{linear} correlation between the two, $R^2=0.92$.}
    \vspace{-5pt}
    \label{fig:linearity_premise}
\end{figure}

In this work, we leverage these semantic properties in order to train regression models.
Our key insight is that, given a latent code, it is possible to accurately predict the magnitude of a semantic attribute in the corresponding image. We do so by measuring its distance from a separating hyperplane \cite{shen2020interpreting} induced by a matching semantic linear latent direction.
\rg{An illustration of such directions is provided in \figref{fig:linearity_premise}(a).}
We show that the latent-space distances can already serve as regression scores for applications where no conventional units are required or exist, \eg ordinal regression tasks. However, we observe that typically these distances are also approximately \textit{linear} with respect to the magnitude of the semantic attribute (see \figref{fig:linearity_premise}(b) for an example). Therefore, they can be calibrated to produce real-world predictions, \eg age in years or head pose in degrees, using linear regression and as few as two labeled samples.
This allows us to perform regression 
in data domains and for semantic attributes where quality supervision is prohibitively difficult to acquire.

\begin{figure}[!hbt]
\setlength{\tabcolsep}{1pt}
    \centering
    \includegraphics[width=0.8\linewidth]{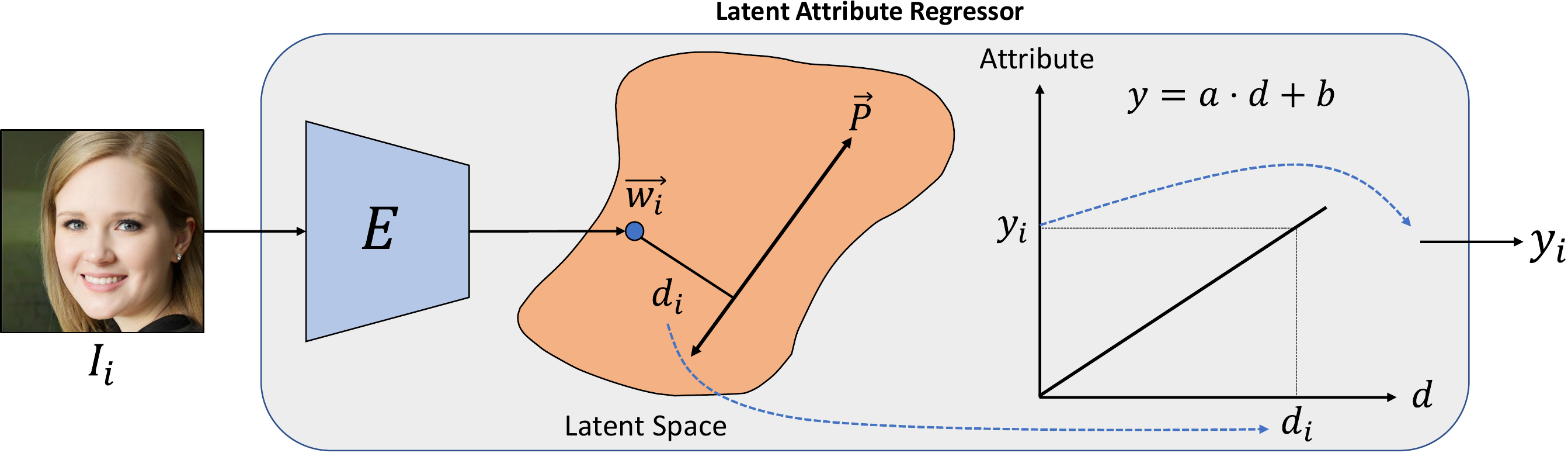}
    \caption{Outline of our proposed regression pipeline. An image $I_i$ is inverted into a latent-code $\vec{w_i}$. The distance $d_i$ of the code $\vec{w_i}$ from a semantic hyperplane $\vec{P}$ is calculated. Finally, $d_i$ is input to the simple regression model which outputs the magnitude of the semantic attribute $y_i$ of the image.}
    \vspace{-9pt}
    \label{fig:pipeline}
\end{figure}

Our model, outlined in \figref{fig:pipeline}, is thus composed of several steps. First, we learn a disentangled, linear, semantic path for an attribute in the latent space of StyleGAN. Finding such paths is a popular research topic in image editing works \cite{xia2021gan}. While \rg{editing} methods differ in the type of supervision and amount of computation required, our approach operates seamlessly with all methods tested. 
This serves as a promising indication that we can leverage any future advancements in editing techniques.

Armed with such a semantic path, we next turn to finding discriminative features which allow us to regress continuous values. We demonstrate that the distance in the latent space from a hyperplane perpendicular to the semantic latent direction is an approximately-linear and smooth representation of the modified property.
However, in order to perform regression on real images, the latent code corresponding to a given image is required. For this end, we show that our method is compatible with off-the-shelf \textit{GAN Inversion} encoders, such as pSp \cite{richardson2020encoding}, e4e \cite{tov2021designing} and Re-Style \cite{alaluf2021restyle}.

Lastly, we observe that the optimal latent space for \textit{GAN inversion} may differ from the optimal space for finding semantic latent directions. In this case, a simple distance from the hyperplane cannot be calculated. 
To bridge the gap, we learn a task-specific mapping between distances in the two spaces using only latent-space considerations, without any additional supervision. This mapping method is outlined in subsection \ref{subsec:bridging}.
By following these steps, our method distills the strength of any semantic property into a single scalar, using only a pretrained generator and weak supervision.

Through extensive evaluation, we show that our model can produce state-of-the-art results on few-shot learning tasks such as pose and age estimation, without \textit{any} direct supervision on other domains, and that it can even match or outperform fully-supervised methods designed for specific tasks and trained on tens of thousands of samples. Where no supervision is available, we show that our model produces meaningful scores by demonstrating its applicability to the task of sorting a collection images by the strength of a semantic property. 

\newpage

In summary, our contributions are:
\begin{itemize}
    \item The observation that commonly, latent-space distances are approximately linearly correlated with the magnitude of semantic properties in an image.
    \item A scheme for converting a pretrained generator and a semantic latent-direction into a state-of-the-art few-shot regressive model.
    \item A new approach to analysing layer-importance and mapping semantic distances between the latent spaces of a GAN.
\end{itemize}

\section{Related Work}

\paragraph{Latent Space of GANs:}
 
Recently, understanding and controlling the latent representation of pretrained GANs has attracted considerable attention. Notably, it has been shown that StyleGAN \cite{karras2019style, karras2020analyzing, Karras2020ada} creates a disentangled, smooth and semantically rich latent space. Many recent works have proposed methods to interpret the semantics encoded in this latent space and apply them to image editing \cite{shen2020interpreting, nitzan2020face, harkonen2020ganspace, abdal2020styleflow, wu2020stylespace, patashnik2021styleclip, spingarn2020gan}. Such methods typically identify a \textbf{linear} latent direction, which when traversed along, starting from an initial latent code, causes a gradual change in a single semantic property of the corresponding generated image.

To perform manipulations on real images, one must first obtain the latent code from which the pretrained GAN can most accurately reconstruct the original input image. This task has been commonly referred to as \textit{GAN Inversion} and has been tackled by numerous recent works \cite{richardson2020encoding, tov2021designing, alaluf2021restyle, zhu2020domain, abdal2019image2stylegan, karras2020analyzing}.
For a thorough introduction to these subjects, we refer the reader to a recent survey \cite{xia2021gan}.

\paragraph{GANs for discriminative tasks:}

Several works sought to leverage the advancement of GANs for discriminative tasks. Arguably, the most straight-forward approach is to simply train a generator and use it to synthesize labeled samples. Such samples are inherently labeled when the generator performs either image-to-image translation between domains or class-conditioned synthesis.
Following this approach, some works achieved competitive results using completely generated datasets \cite{tanaka2019data, marriott2020assessment}. Others enriched a real dataset to improve performance in the low-data domain \cite{zhu2017data, frid2018gan, antoniou2017data, wang2018low} or on biased or unbalanced data \cite{mariani2018bagan, hochberg2021style, ramaswamy2020fair, grover2019fair, sharmanska2020contrastive}.
When applied to the ImageNet classification task \cite{imagenet_cvpr09}, some works \cite{ravuri2019classification, shmelkov2018good} demonstrated that performance sees only modest improvement when enriching the real set, and becomes poor when replacing it.
Deviating from labeled-image generation, some works \cite{springenberg2015unsupervised, odena2016semi, salimans2016improved} proposed to use GANs for semi-supervised learning, where the classifier additionally serves as a discriminator. Others \cite{ratner2017learning, sixt2018rendergan, perez2017effectiveness} proposed to learn parameters and compositions of simple image augmentations for existing data.

While these works cover a diverse collection of settings and approaches, they all used GANs in order to eventually generate more data \textbf{for training}.
In the context of discriminative tasks, several recent methods have proposed to utilize GANs for additional purposes. 
Lang \etal \cite{lang2021explaining} used StyleGAN \cite{karras2020analyzing} to visualize counterfactual examples for explaining a pretrained classifier's predictions. Chai \etal \cite{chai2021ensembling} used style-mixing in the fine-layers of StyleGAN to generate augmentations that are ensembled together at test-time.
Most related to our work is the representation learning framework GHFeat \cite{xu2020generative}. In their work, Xu \etal train an encoder for \textit{GAN Inversion} into the latent space of a pretrained StyleGAN and demonstrate that the visual features learned by this encoder can be used to train a variety of models for downstream tasks in a fully supervised manner.

In contrast, we build on recent progress in the study of GANs' latent space and observe that distances within this space can already serve as \rg{one-dimensional discriminative features}. Our method can leverage these features for simple discriminative tasks, such as sorting, using only weak supervision. With just a few labeled samples, as little as two, we are able to regress real world values. In settings where both our models are applicable, we compare our method with GHFeat and find that we can obtain more accurate results using significantly less data.

\paragraph{Few-shot regression:}

Few-shot learning refers to the task of training a model to reason over a specific task, provided with only a few labeled training samples.
In the context of visual media, such methods have been widely studied for classification \cite{snell2017prototypical,vinyals2016matching,sung2018learning}, object detection \cite{wu2020multi,Fan2020FewShotOD} and segmentation \cite{endo2021few,rakelly2018few}. Relatively few image-related few-shot models have stepped beyond these bounds. They tackle image-to-image translation \cite{liu2019few}, super resolution \cite{soh2020meta}, motion prediction \cite{gui2018few}, and re-identification \cite{wu2019few}.

In particular, this scarcity holds for regression tasks, where a majority of research focused on pose estimation using strong task-specific priors \cite{xiao2020few} or applying meta-learning ideas to keypoint extraction and gaze estimation frameworks \cite{tseng2019few, park2019few}. For a thorough overview, we refer the reader to a recent survey \cite{wang2020generalizing}. These works, however, all deal with specific tasks and adapt them across domains in a few-shot manner. Their use of task-specific supervision, priors and architectures prevents them from easily being extended to new objectives. In contrast, our method can be easily generalized across domains and different regression goals, including cases where collecting task-specific supervision from other domains would be difficult or outright impossible.

\section{Method}

A shared premise in GAN-based editing works \cite{shen2020interpreting, harkonen2020ganspace, patashnik2021styleclip} is that single, semantic image properties can be manipulated through modifications of the latent codes used to generate the image.
These modifications are conducted by discovering appropriate \textbf{global}, \textbf{linear} steering directions within the latent space of the generator \rg{(see \figref{fig:linearity_premise}(a).} Such directions take the form of vectors, $\vec{n}$, which can be used to induce a semantic change in \textit{any} code $\vec{w}$ (\ie they are \textbf{global}) through \textbf{linear} addition: $\vec{w}' = \vec{w} + \alpha\vec{n}$, where $\alpha$ is a scalar that controls the strength of the modification.

Studying these latent directions, we make an important observation -- not only are they linear, but oftentimes the magnitude of their effect is linear with respect to $\alpha$. Traversing one unit of distance across $\vec{n}$ causes a roughly fixed-size change in the property in question (\figref{fig:linearity_premise}(b)). In our work, we rely on this observation and use the distance along a linear latent semantic direction to regress a corresponding property in a given image. 

To regress a single property in a real image $I_i$, we thus require two components: a semantic latent direction corresponding to the property in question $\vec{n}$, and the representation of the image in the latent-space of a pretrained GAN: $\vec{w_i}$. As previously discussed, semantic directions can be discovered using an array of editing methods \cite{shen2020interpreting, shen2020closedform, patashnik2021styleclip}.
An alternative approach to considering such directions is to view them as normals to a hyperplane
$\vec{P}$ partitioning the latent space: $\vec{n}\cdot\vec{w} + b = 0$. Now, the magnitude of the property in the image is the distance of its latent representation to the hyperplane:
\begin{equation}  \label{eq:dist_eq}
    d = dist(\vec{w}_i, \vec{P}) =  \vec{w}_i \cdot \vec{n} + b.
\end{equation}
\yn{Some editing methods produce a value for the intercept $b$, while others do not.}
When $b$ is unknown, we set it arbitrarily to $b=0$. As $b$ modifies all distances by a constant factor, dropping it has no effect on any sorting applications, and it can be effectively determined during calibration when training a linear regressor on real data. 
\yn{When $b$ is known, the distance of 0 carries a special semantic meaning. For example, when considering head pose it corresponds to a frontal face}.

While calculating the distance is a simple algebraic operation and requires no supervision, this does not imply that the method is unsupervised. Supervision is dictated by the methods used for finding the semantic latent direction and for performing GAN Inversion. These methods typically require weak or indirect supervision. InterFaceGAN \cite{shen2020interpreting}, for example, requires only binary labeling of an attribute. In the head pose example, such annotation would amount to left/right pose information, rather than an explicit yaw angle, which is significantly harder to annotate. The GAN itself is trained in a completely unsupervised manner.

\subsection{Inversion}
\label{subsec:inversion}

A requirement of our approach is the ability to invert an image into the latent space of the GAN. Specifically, with StyleGAN, there exist numerous latent spaces that have been considered by previous works. We follow the common approach of inverting an image into the \wplus space, first introduced by Abdal \etal \cite{abdal2019image2stylegan}. 

For an inversion method, we follow the encoder-based scheme. Our method works seamlessly with multiple off-the-shelf encoders, but shows improved performance when utilizing e4e \cite{tov2021designing}, which was designed to produce latent codes that are highly editable. \nuke{We provide a comparison of inversion methods in the appendix.}

\subsection{Bridging the Gap Between Spaces}
\label{subsec:bridging}
A topic of ongoing research in the field of GAN-based image manipulation is the choice of optimal latent spaces for inversion \cite{tov2021designing} and editing \cite{wu2020stylespace}. In StyleGAN \cite{karras2020analyzing}, the generator supports four \nuke{commonly used} latent spaces: $\mathcal{Z}$, \w, \wplus and $\mathcal{S}$. 
$\mathcal{Z}$ denotes the space of initial codes, sampled from a Gaussian prior, while the others denote increasingly complex sets of codes obtained by passing through StyleGAN's mapping network (\w), by assigning different \w codes to different layers of the GAN (\wplus) or by applying affine transformations to the codes ($\mathcal{S}$). These different spaces have been shown to possess different attributes, ranging from their ability to support accurate reconstructions of images, to their linearity and ability to support disentangled editing operations.

A common challenge in StyleGAN editing tasks is that the latent space most often used to identify semantic latent directions, \w, is not expressive enough to support accurate reconstructions of images. The \wplus space, meanwhile, allows for more accurate reconstructions - but does not behave well under \w-based transformations. One reason for this behaviour is that in the \wplus space, different codes affect different layers which in turn affect different semantic properties of the generated image. For example, pose is controlled by early layers of the network while colors are largely controlled by later layers. While applying a \w space editing direction equally to all layer codes may still modify a desired property, not all layer modifications are required, or even affect the property at all. A natural question arising in such a case is then - which layers are relevant to the property at hand? We propose to answer this question by learning an importance score for each layer in an unsupervised manner. 

To do so, we sample a random latent code $w \in \mathcal{W}$ in the same space as the semantic hyperplane. We edit the sampled code and obtain $w_{e}$ which is then used to generate an edited image $I_{e} = G \left(w_{e}\right)$. We map $w$ to a code $w^+ \in \mathcal{W}+$ by duplicating it, once for each layer. Finally, we set up a direct optimization scheme where we attempt to modify the mapped code, $w^+$ such that it can be used to generate the edited image, i.e. we aim to solve:
\begin{equation} \label{eq:optimize_w}
    w^* = argmin_{w^+} \left\|G\left(w^+\right) - I_{e}\right\|^2~.
\end{equation}
By solving \eqnref{eq:optimize_w} using gradient descent, we can track the magnitude of gradients along different entries of the latent code. Entries with greater change are more significantly tied to the modified semantic property.
We repeat this process with multiple codes and compute the mean over the discovered gradient directions, thus obtaining an importance weighting for the given semantic over the layers. In order to avoid spurious results due to different layers having different gradient scales, we normalize gradients by their values when optimizing between unrelated-images.
Armed with per-layer importance scores $\{S_i\}_{i=1}^{L}$, we use them to calculate a weighted sum of hyperplane distances:
\begin{equation}
    d_{\mathcal{W}+} = dist_{\mathcal{W}+}(\vec{w}^+, \vec{P}_{\mathcal{W}}) = \sum\limits_{i = 1}^{L}{S_{i} \cdot dist_\mathcal{W}(\vec{w}^+_i, \vec{P}_{\mathcal{W}})}~,
\end{equation}
which we use as an effective distance between $\vec{w}^+ \in \mathcal{W}+$ and a hyperplane $\vec{P}_{\mathcal{W}}$ given in \w. 

\subsection{Calibration}
\label{subsec:calibration}

In many cases, one expects a regression model to output a prediction in ``real-world" values, such as head pose in degrees. As presented, the latent-distances are uncalibrated. One can ask, given a sample for which $d=3$, what is the actual head pose \nuke{in degrees}?
At this point, we turn to our prior observation: the magnitude of change in a property is roughly a linear function of the distance.
Hence, to calibrate the distances to actual real-world values, we need only train a simple linear regression model with one feature per sample - the distance to the hyperplane. Such a linear model simply takes the form $y = a\cdot d + b$, where $y$ is the calibrated property prediction, $d$ the latent-space distance, and $a$, $b$ are learned parameters which can be determined with as few as two sampled points.

Once trained, the model gets the distance from the hyperplane as input and predicts the real-world interpretable value. Finally, the entire pipeline produces a few-shot image-property regression model.

\section{Experiments}

We provide evaluations for several domains and properties. When ground truths are not available, we either annotate existing datasets using pretrained networks (WHENet \cite{zhou2020whenet} and DEX \cite{Rothe-ICCVW-2015}) or evaluate our performance by considering relative ordering of image collections.
Human face image experiments use the official StyleGAN2 pretrained on FFHQ \cite{karras2019style}. Age experiments use the CACD \cite{chen14cross} and CelebA-HQ \cite{karras2017progressive} datasets, while all other evaluations are conducted on CelebA-HQ. Leaf image experiments use the Plant-Village dataset \cite{hughes2015open}.
Cat image experiments use the official StyleGAN-ADA \cite{Karras2020ada} model trained on AFHQ \cite{choi2020starganv2}, and are evaluated on the test-split of the same set. 
Experiments on additional domains are provided in the appendix. In all cases, the GAN was trained in a completely unsupervised manner, without any labels. All results are shown on real (\ie non-generated) images.

For methods relying on calibration, including ours and all baselines, we evaluate performance by training a thousand times over randomly sampled subsets of the data. We report the mean and standard deviation for each experiment.

\subsection{Feature space comparisons}
\label{subsec:feature_space}

We start by demonstrating that distances in the latent space of the GAN are more semantically meaningful and better behaved than equivalent distances in alternative feature spaces.
In particular, we compare our method with four baselines operating in a similar manner to the InterFaceGAN \cite{shen2020interpreting} approach: First, a large collection of binary-tagged images are acquired, along with their representation in each chosen feature space. Then, we train an SVM in the given feature space using the binary labels, providing us with a separating hyperplane matching the semantic attribute described by the labels. Finally, the distance to the hyperplane within the given feature space is used as a discriminative feature for training a linear regressor.

The simplest baseline is \textit{Pixel SVM} which considers the pixel space of the image as the \nuke{given} feature space. Notably, this space doesn't contain any explicit semantics and suffers from high-dimensionality, \ie it may be prone to overfitting the data. In order to mitigate this issue, we propose a second baseline, \textit{PCA SVM}, which applies PCA to the images and represents them using just the first $30$ principal components. A third baseline, \textit{Deep SVM}, operates in the feature space learned by a ResNet18 \cite{he2016deep} trained to provide binary classification of the images. \workwork{Lastly, GHFeat \cite{xu2020generative} demonstrated that style-space (\s) codes extracted using a \textit{GAN Inversion} encoder are useful features for discriminative tasks.} We devise the corresponding \textit{GHFeat SVM} baseline which operates in the feature space extracted by a pretrained GHFeat encoder. In order to provide visual reference, we also compare with a method that always predicts the mean value of the attribute over the training set.

\begin{figure}[h!]
\setlength{\tabcolsep}{1pt}
    \centering
    \begin{tabular}{c c}
        \includegraphics[width=0.5\linewidth]{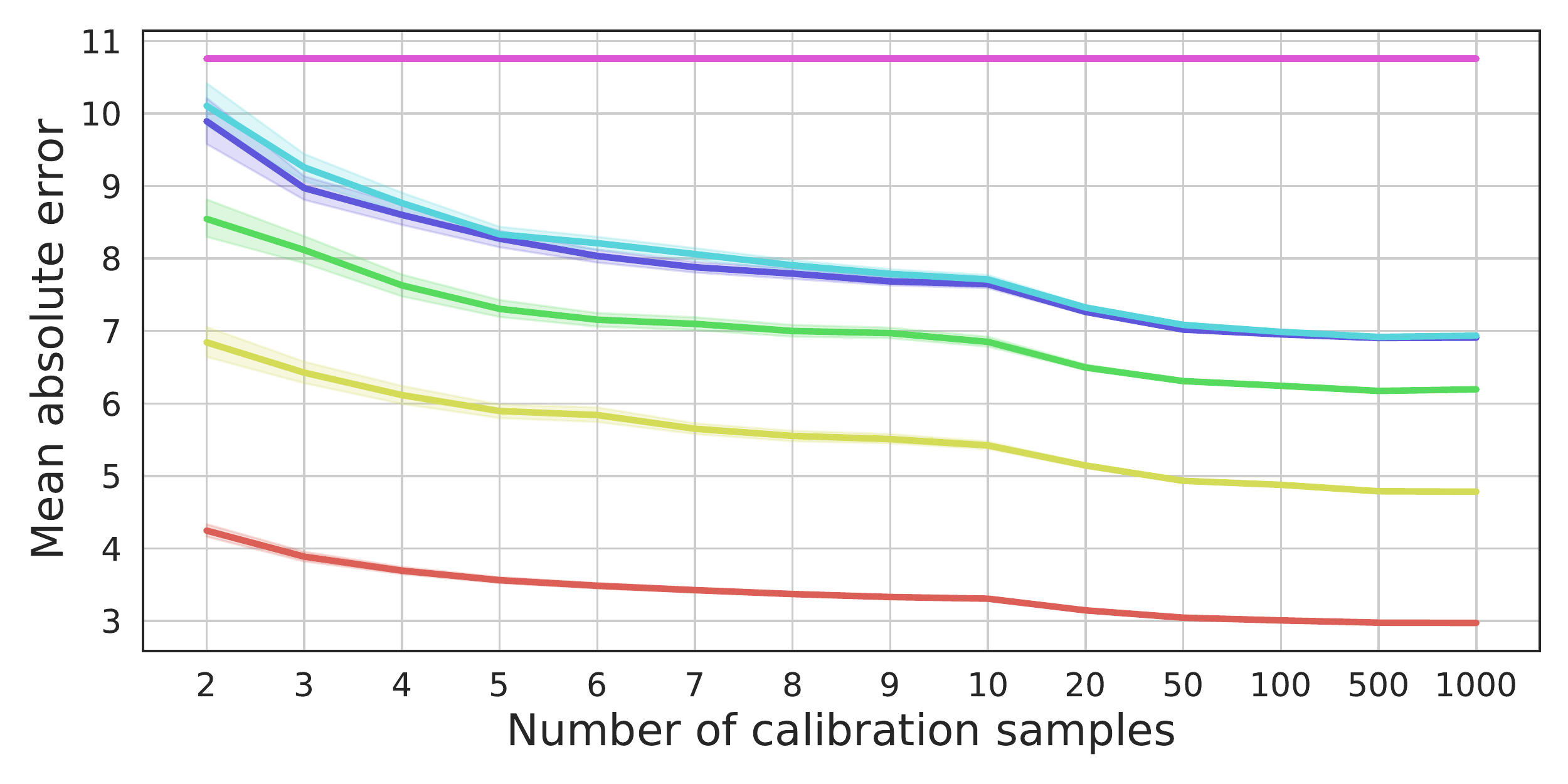} &
        \includegraphics[width=0.5\linewidth]{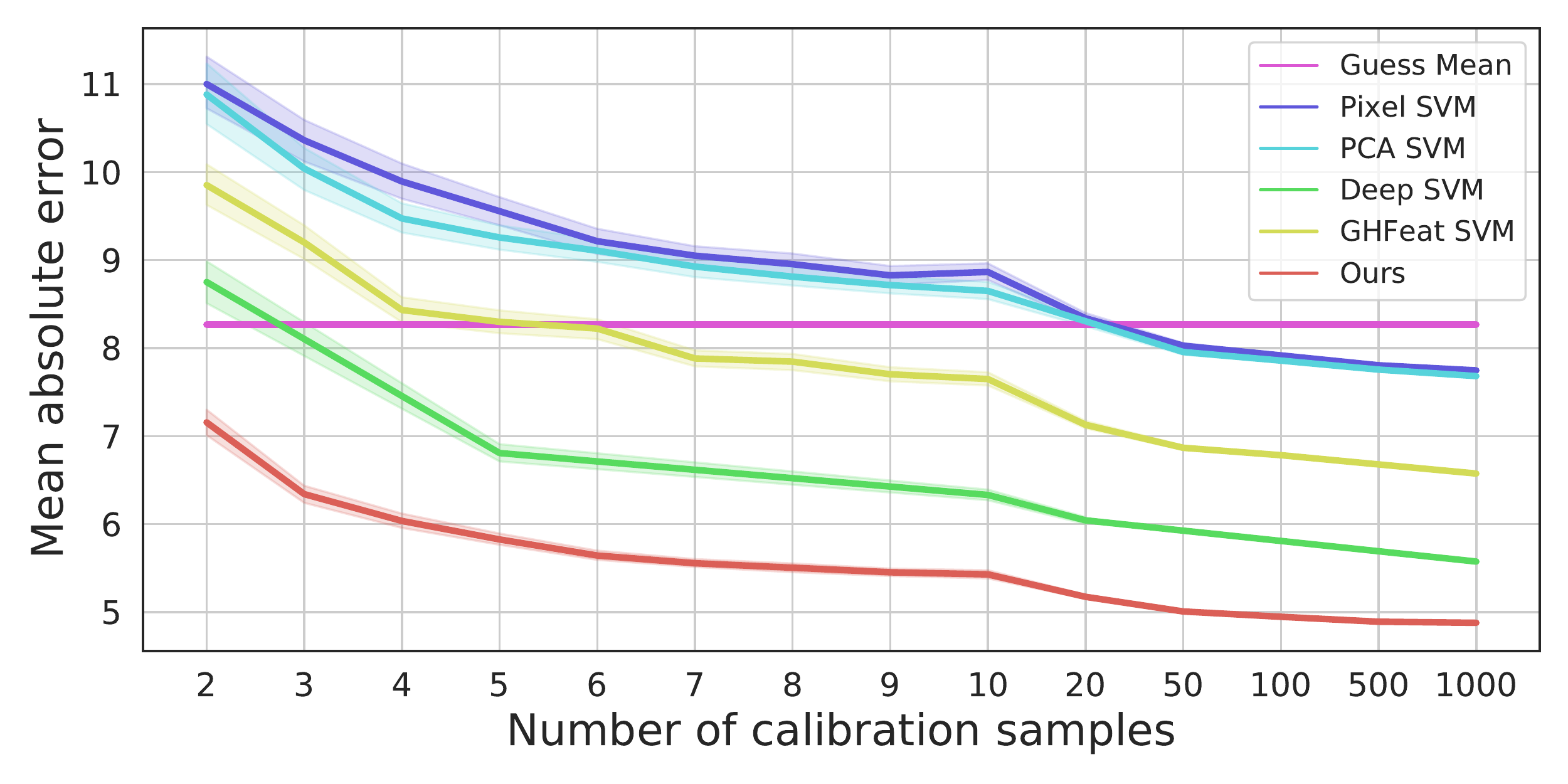} \\
        (a) Pose & (b) Age
    \end{tabular}
    \caption{
    Quantitative comparisons to baselines as a function of the number of labeled images used for calibration. Results are provided on the CelebA-HQ dataset \cite{karras2017progressive}. All methods operate in a similar manner on different feature spaces.
    Our method outperforms all baselines by a large margin.
    }
    \label{fig:comparison_to_svm_baselines}
\end{figure}

As can be seen in \figref{fig:comparison_to_svm_baselines}, for all experiments, our model outperforms the baselines, showing that distances to semantic boundaries within the latent space of a pretrained generator are more semantically meaningful, and serve as better discriminative features for \textit{linear} regression than distances in alternative feature spaces. 
Furthermore, these results demonstrate that the idea of utilizing a one-dimensional distance metric in some learned feature space is not universal, but relies upon the extensive semantic knowledge encapsulated by the GAN. Indeed, for some configurations, the baselines perform worse than simply returning the mean of the training set's distribution.


\subsection{Evaluating linearity}

\begin{figure}[h!]
\setlength{\tabcolsep}{1pt}
    \centering
    \begin{tabular}{c c}
        \includegraphics[width=0.5\linewidth]{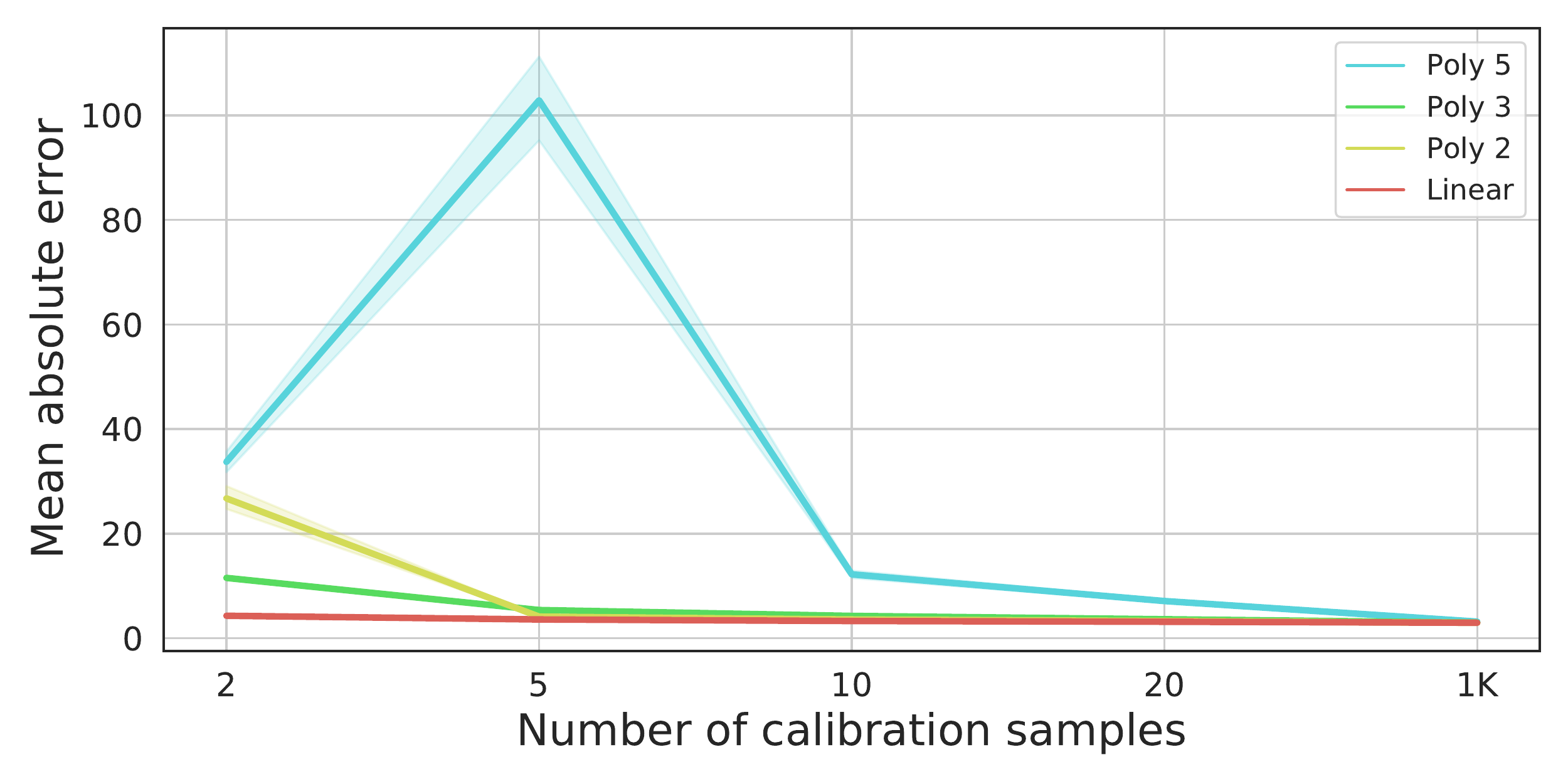} & \includegraphics[width=0.5\linewidth]{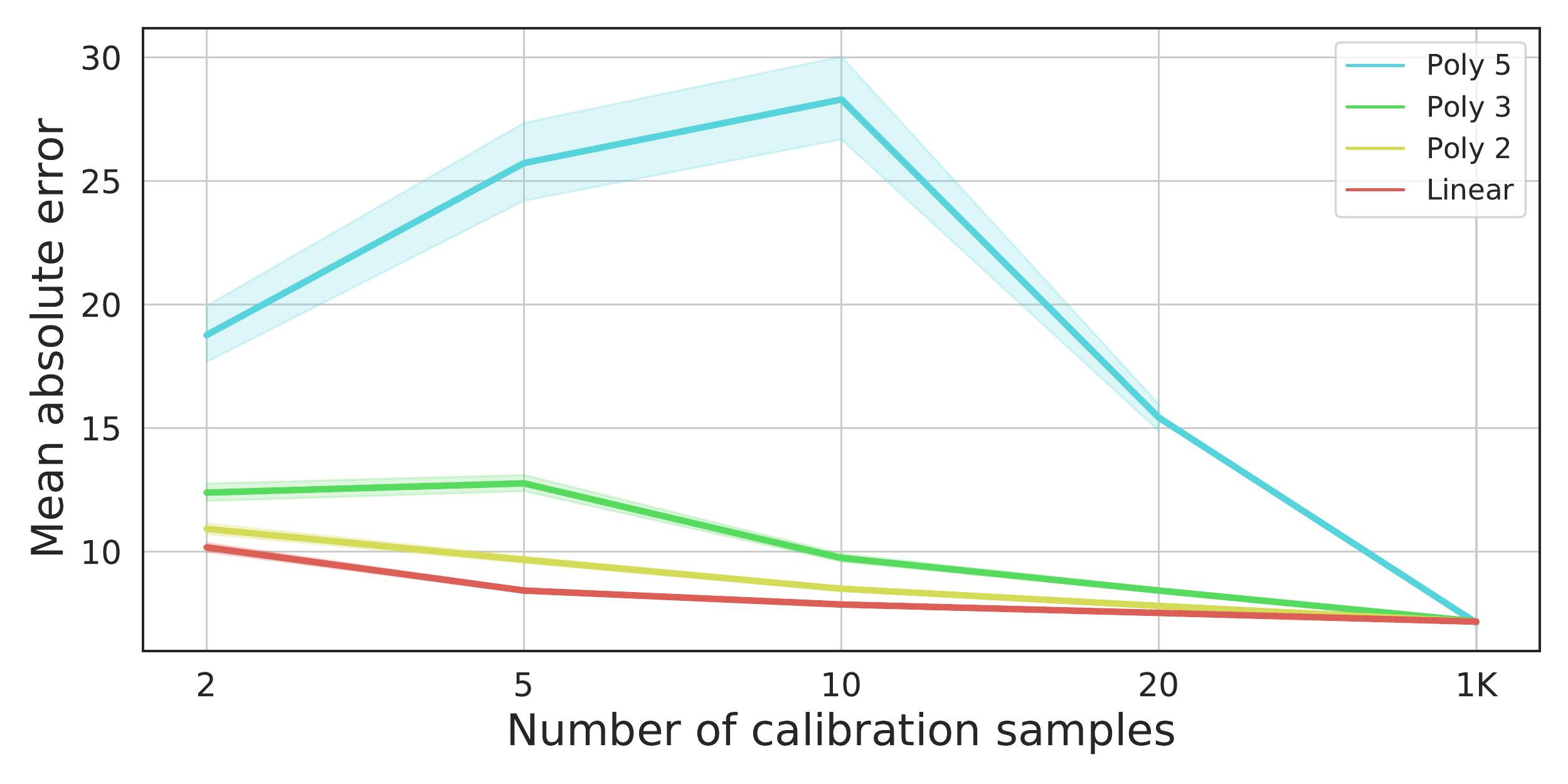} \\
        (a) Pose & (b) Age
    \end{tabular}
    \caption{
    Quantitative comparison of four different choices for the degree of fitted regression functions. In addition to the linear model outlined in our method, we evaluate polynomial degrees of $2$, $3$ and $5$. The linear model outperforms the alternatives when only a few samples are available, and is equivalent for a thousand samples. In (a) pose is evalutated on CelebA-HQ \cite{karras2017progressive} and in (b) age is evaluated on CACD \cite{chen14cross}.
    }
    \vspace{-5pt}
    \label{fig:comparing_polynomial_degrees}
\end{figure}

We provide a dedicated experiment demonstrating the approximately linear correlation of the attribute's magnitude and the distance-features. 
Using the same set of labeled images and their distances from a semantic hyperplane, we compare the accuracy of a linear model trained on these data to the accuracy of polynomial models of higher degrees - specifically $2$, $3$ and $5$. Results are shown in \figref{fig:comparing_polynomial_degrees}. 
As can be seen, for all attributes tested, the linear model is superior when a few labeled samples are provided. Additionally, despite having greater expressive power, the polynomial models do not outpace the linear model even when a thousand labeled samples are provided.

The ability of a linear model to more accurately capture the correlation between the distance-feature and the magnitude of the attribute, implies that it is indeed approximately linear.

\subsection{Calibrated results}
\label{subsec:calibrated_results}

Having established that the latent space of StyleGAN is a source of semantically meaningful feature representations, we turn to evaluating the strength of these features with respect to prior works.
Towards this end, we demonstrate the performance of our model on a set of regression tasks: pose and age estimation for human faces. Additionally, pose estimation for cars is provided in the appendix. We train and compare models over a wide range of supervision settings, showing that our approach achieves state-of-the-art performance in the few-shot domain as well as surprisingly competitive results on larger datasets, even when compared to methods designed for a single, specific task.

For each attribute we evaluate against a set of methods designed for the specific task, often utilizing a task specific architecture. Furthermore, we compare against GHFeat \cite{xu2020generative}, which proposed the use of a different set of GAN-inspired features, and is thus comparable in all scenarios. Comparisons against GHFeat are conducted by training a linear regression model directly on their features. Additionally, we compare against the \textit{GHFeat SVM} variant introduced in \subsecref{subsec:feature_space}, which is more similar to our approach, and which we find to perform better than the original in few-shot scenarios.

\begin{figure}[h!]
\setlength{\tabcolsep}{1pt}
    \centering
    \begin{tabular}{c c}
        \includegraphics[width=0.5\linewidth]{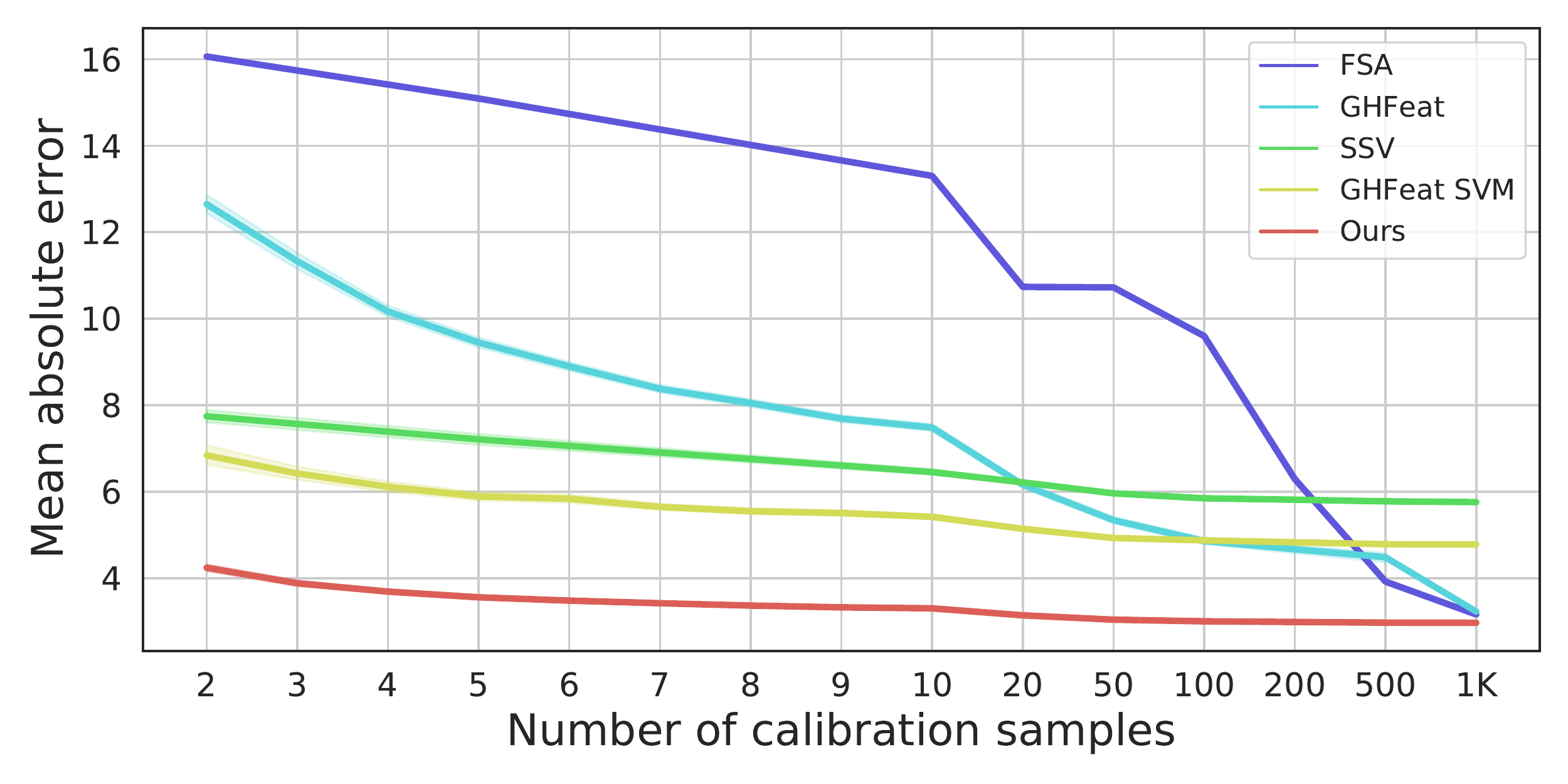} & \includegraphics[width=0.5\linewidth]{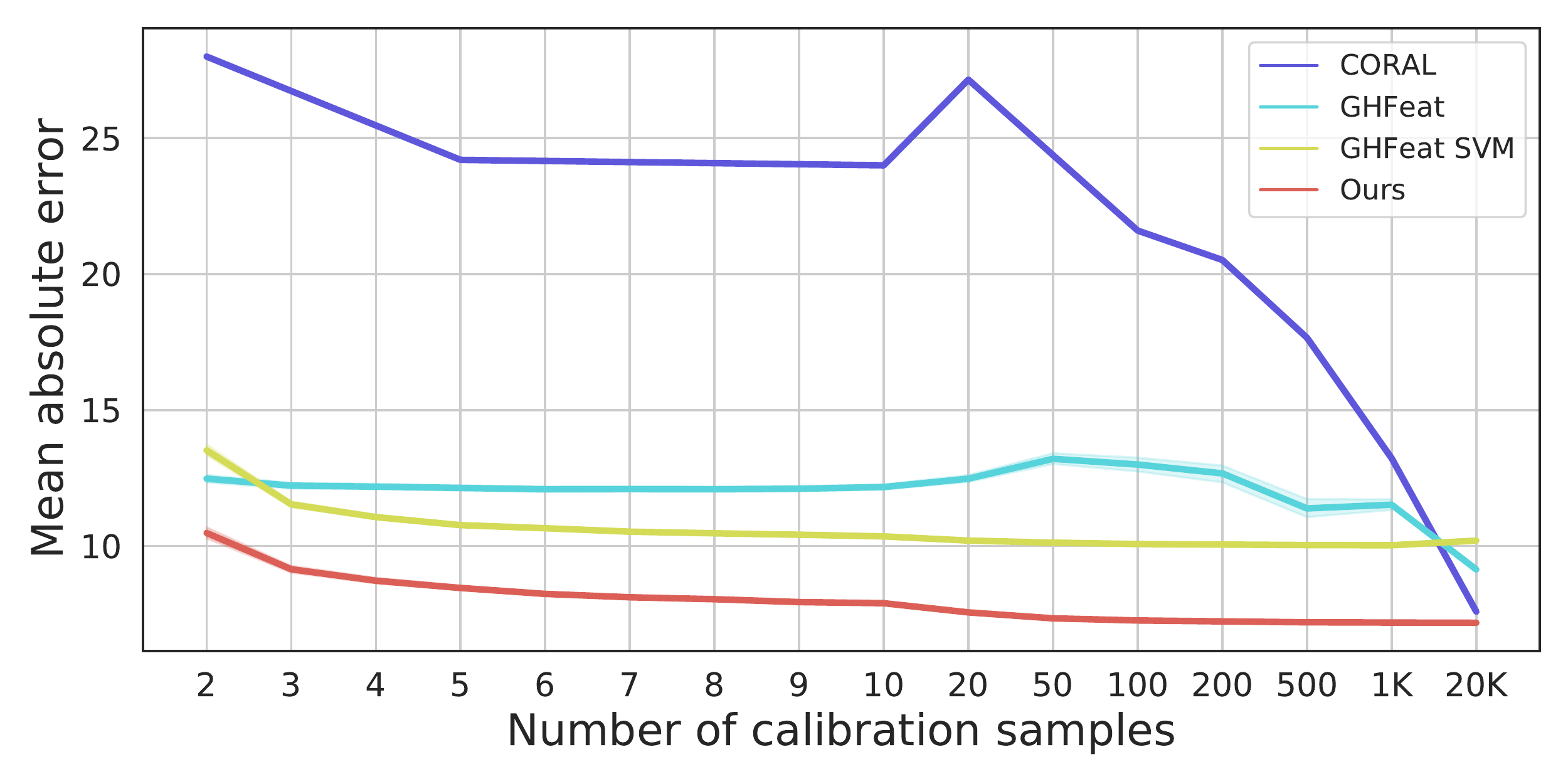} \\
        (a) Pose & (b) Age
    \end{tabular}
    \caption{
    Comparisons to alternative models as a function of the number of labeled images used in training. In (a) we compare to GHFeat \cite{xu2020generative}, FSA \cite{yang2019fsa} and SSV \cite{mustikovela2020self} on the CelebA-HQ dataset \cite{karras2017progressive}. In (b) we compare to GHFeat and CORAL \cite{cao2020rank} on the CACD dataset \cite{chen14cross}.
    }
    \vspace{-5pt}
    \label{fig:comparison_to_methods}
\end{figure}

\paragraph{Human Pose:} We evaluate the performance of our model on the task of predicting human head poses, and specifically yaw. The semantic direction used by our method was extracted using InterFaceGAN \cite{shen2020interpreting}. In addition to GHFeat, we compare to two pose estimation methods: FSA \cite{yang2019fsa}, a fully supervised method and SSV \cite{mustikovela2020self}, a method which learns pose estimations in a self-supervised manner and then calibrates them to dataset-specific values with few labeled samples. In this sense, their method can also be regarded as a few-shot approach. In \figref{fig:comparison_to_methods}(a) we show the mean absolute errors (MAE) of yaw estimation using the outlined approaches. Our method consistently outperforms all methods when presented with limited supervision, and remains competitive up to a thousand labeled samples. Of particular note is the fact that our model displays better performance than SSV, indicating that the latent space of a pretrained GAN encodes more meaningful pose information than comparable self-supervised methods which were designed and trained specifically to extract pose.

\paragraph{Human Age:}
We evaluate our performance on the task of human age estimation. The age editing direction used by our method was discovered through natural language descriptions using StyleCLIP \cite{patashnik2021styleclip}. Specifically, the boundary was extracted with the prompts ``old face" and ``young face".
In addition to GHFeat, we compare to CORAL \cite{cao2020rank}. \figref{fig:comparison_to_methods}(b) shows the MAE on age estimation in years. Our method outperforms the alternatives when presented with limited data, and is only outpaced by CORAL when provided with tens of thousands of samples. Specifically, our 20-sample model obtains a $\text{MAE}$ of $ 7.46$, in line with CORAL's 20K-sample result -- $7.59$. Lastly, this experiment demonstrates that textually described boundaries are suitable for regression - opening a path to few-shot regression of many properties that can be reasonably described through natural language.

As verified by our experiments, our method achieves state-of-the-art results in the few-shot domain and in some scenarios remains competitive even when compared to models trained on a thousand times as many images. Our advantage holds even when compared to models designed for specific regression tasks. Furthermore, our method operates across multiple domains and can be easily adapted to work with a range of latent-space editing approaches. Lastly, our results indicate that the relationship between the latent space distance-features and real-world property values can indeed be well approximated through a \textit{linear} mapping.

\subsection{Uncalibrated results}

We next discuss domains and attributes for which continuous supervision is not available. Even in such cases, our method can still produce meaningful scores which describe the magnitude of a semantic property in the image. With no supervision given, the scores cannot be calibrated to any real-world human-interpretable value. Nevertheless, such scores are still useful for applications such as sorting and ordinal regression. Here, we demonstrate their applicability to the task of sorting a collection of images by a given property -- \eg according to how happy the person in the image is.

\begin{figure}[!hbt]
    \centering
        
    \includegraphics[width=0.9\linewidth]{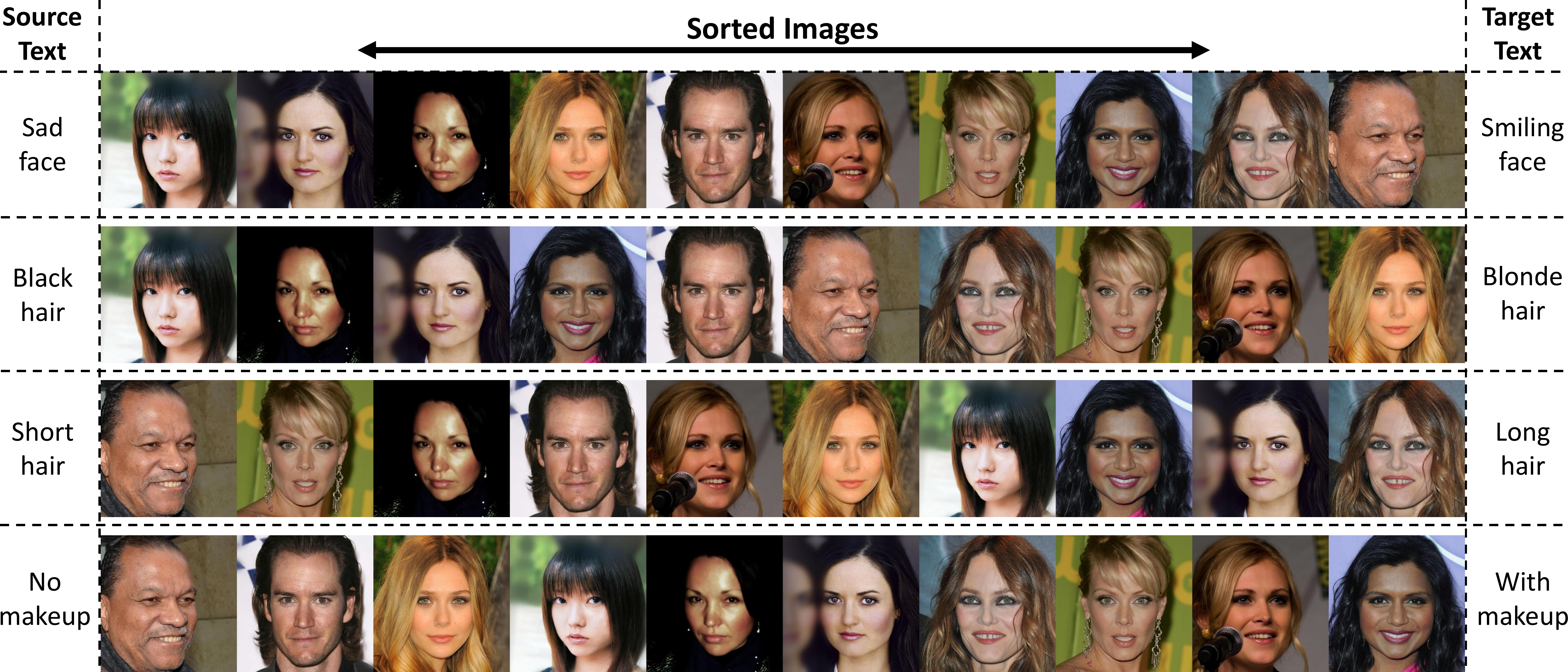}
    \caption{Sorting images according to textual descriptions of semantic properties. In each row, we sort the same set of randomly sampled CelebA-HQ \cite{karras2017progressive} images according to their distance from a text-based editing boundary extracted by StyleCLIP \cite{patashnik2021styleclip}. Each row's editing direction is induced by the source text (left) and the target text (right).}
    \vspace{-3pt}
    \label{fig:clip_sort}
\end{figure}

In \figref{fig:clip_sort} we show results for sorting a set of face images according to properties described through textual prompts, using distances to boundaries discovered through 
StyleCLIP's global mapping approach. We sort the same randomly sampled collection according to four distinct attributes: expression, hair color, hair length, and amount of makeup.

\begin{figure}[!hbt]
\setlength{\tabcolsep}{1pt}
    \centering
    \begin{tabular}{l}
        \hspace{0.15cm} Sicker \hspace{8.55cm} Healthier \\
        \includegraphics[width=0.8\linewidth]{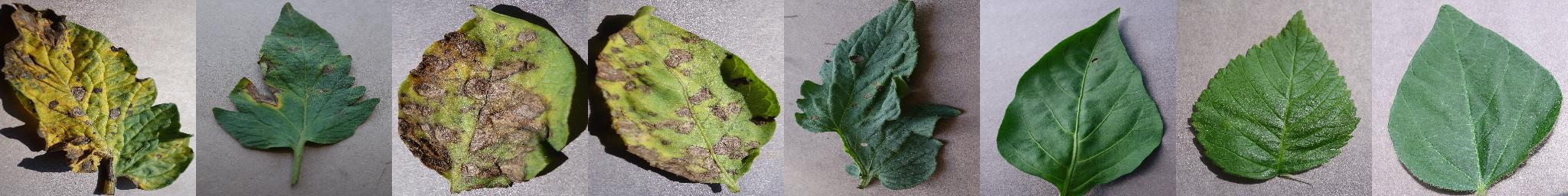} \\
        \includegraphics[width=0.8\linewidth]{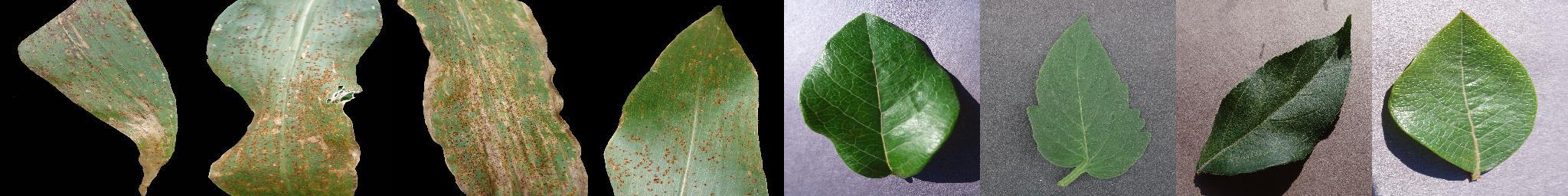} \\
        \includegraphics[width=0.8\linewidth]{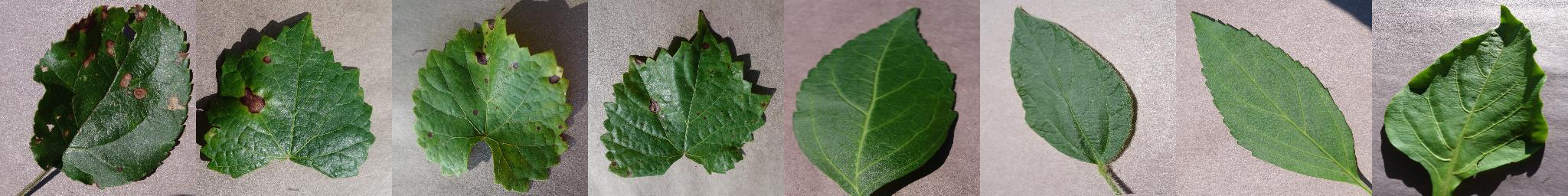} \\
    \end{tabular}
    \caption{Sorting images from the Plant-Village \cite{hughes2015open} dataset using a semantic direction extracted by InterFaceGAN \cite{shen2020interpreting}. In each row, we sort randomly sampled sets containing images labeled as either healthy or sick. We separate rows by type of disease to facilitate visual comparisons.
    }
    \vspace{-5pt}
    \label{fig:leaves_sort}
\end{figure}

In \figref{fig:leaves_sort} we show the results of sorting a collection of leaf images corresponding to a 'sick or healthy' direction, using a boundary extracted with InterFaceGAN \cite{shen2020interpreting} based on binary sick / healthy annotations. Our method successfully turns these binary annotations into continuous values which allow us to determine which leaves are sicker than others. For example, in the last row, the number of ``black spots" decreases gradually while moving from the most-sick leave towards the healthy leaves.

\begin{figure}[!hbt]
\setlength{\tabcolsep}{1pt}
    \centering
    \begin{tabular}{c c}
        \raisebox{0.04\textwidth}{Yaw} & \includegraphics[width=0.85\linewidth]{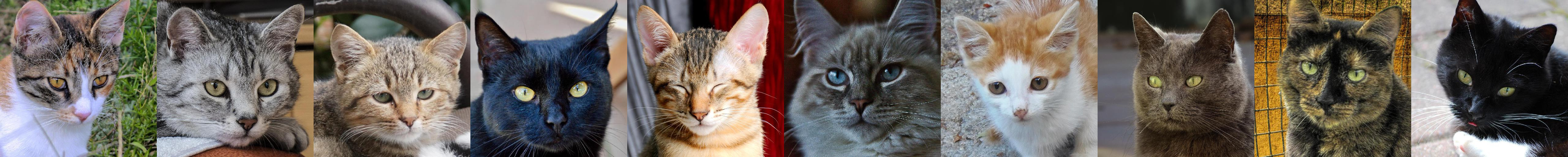} \\
        \raisebox{0.04\textwidth}{Age} & \includegraphics[width=0.85\linewidth]{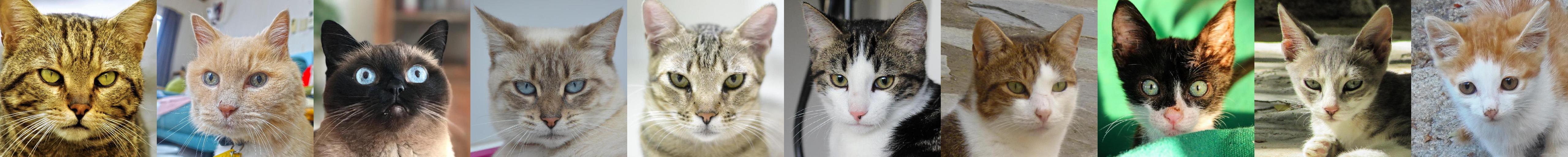} \\
        \raisebox{0.04\textwidth}{Pitch} & \includegraphics[width=0.85\linewidth]{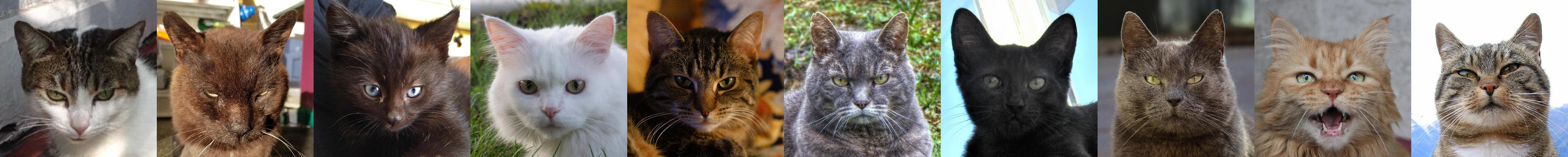} \\
    \end{tabular}
    \caption{Sorting images from AFHQ-cat \cite{choi2020starganv2} using semantic directions extracted by SeFA \cite{shen2020closedform}. }
    \vspace{-7pt}
    \label{fig:sort_afhq_cats}
\end{figure}

In \figref{fig:sort_afhq_cats} we show results on sorting collections of cats randomly sampled from the AFHQ dataset \cite{choi2020starganv2}, using semantic directions discovered in an unsupervised manner with SeFA \cite{shen2020closedform}. Our method extends seamlessly to these additional domains and semantic direction discovery approaches.

As can be seen, the ordered results largely align with human expectation. In order to verify this claim we conduct a user study on the human and cat face domains. The full details of the baselines and the experimental setup are provided in the appendix. Results are summarized in \tblref{tbl:sort_user_study}. As verified by the study, our model learns to regress more consistent scores for images across both domains.
In particular, by augmenting the semantic understanding of CLIP with the knowledge embedded in the latent space of the pretrained GAN, we are able to improve results over a CLIP-only-baseline.

As demonstrated through our experiments, our approach learns to regress meaningful, uncalibrated values across multiple domains and using a wide range of latent-direction discovery methods.

\vspace{-5pt}
\begin{table}[!hbt]
    \centering
    \caption{User study results for sorting quality. For each \yn{attribute} we report the percent of responders which preferred the sorting induced by each of three sorting methods. Our method consistently provides an order which is more consistent with human preference. See the appendix for more details.}
    \begin{tabular}{c c}
    
    \begin{tabular}{l c c c}
        Attribute & Ours & CLIP & Random \\ \hline
        Hair color & $\mathbf{73.55\%}$ & $25.81\%$ & $0.65\%$ \\ 
        Makeup & $\mathbf{70.53\%}$ & $18.25\%$ & $11.23\%$ \\ 
        Expression & $\mathbf{53.45\%}$ & $43.27\%$ & $3.27\%$ \\ 
        Hair length & $\mathbf{84.73\%}$ & $2.18\%$ & $13.09\%$ \\ \hline
        Average & $\mathbf{70.56\%}$ & $22.38\%$ & $7.06\%$
    \end{tabular}
    &
    \begin{tabular}{l c c c}
        Attribute & Ours & SSV & Random \\ \hline
        Yaw & $\mathbf{66.67\%}$ & $28.99\%$ & $4.35\%$ \\ 
        Pitch & $\mathbf{82.50\%}$ & $7.50\%$ & $10.00\%$ \\ \hline
        Average & $\mathbf{75.71\%}$ & $16.71\%$ & $7.58\%$
    \end{tabular}
    \\
    (a) Human faces & (b) Cats
    \end{tabular}
    \vspace{-5pt}
    \label{tbl:sort_user_study}
\end{table}

\section{Discussion}

We have presented a method for leveraging implicit knowledge of a generative model for regression tasks.
Our key take home message is that, through unsupervised training, GANs learn to encapsulate detailed semantic knowledge within their latent space. This knowledge can be used to provide invaluable supervision for downstream tasks.
We demonstrated that distances traveled along semantic latent directions are highly descriptive features and are typically linearly correlated with the magnitude of change in semantic attributes of images.
These observations therefore allow us to train regression models for a multitude of attributes using only weak forms of supervision and as few as two labeled samples. These models achieve state of the art results in the few-shot domain.  

Our work focused on the current state of the art in general image synthesis - StyleGAN. However, \yn{it} is not the only generative network to exhibit an editable latent space. Alternative networks such as BigGAN similarly support editing operations \cite{jahanian2019steerability, spingarn2020gan}, and may allow for regression on attributes or datasets which StyleGAN struggles with.

While we have observed linearity in all semantic attributes which we tested, this property is unlikely to hold universally. Indeed, for some attributes, even finding a disentangled latent direction is infeasible.
Furthermore, similar to other methods which rely on StyleGAN \cite{lang2021explaining, chai2021ensembling}, our method obtains better results when operating on images within the domain used to train the GAN.
This limitation stems in part from to the inability of current \textit{GAN Inversion} methods to reconstruct out-of-domain images while preserving latent semantics. 
However, as this semantic-preserving out-of-domain inversion task is of interest to the community at large, we are looking forward to see this barrier overcome.

We hope that our work can inspire others to consider the latent space of GANs as a source of semantically-rich supervision which can be leveraged to tackle a wide range of downstream tasks.

{\small
\bibliographystyle{ieee}
\bibliography{egbib}
}

\appendix
\appendixpage
\section{Broader Impact}
Our model consists of a general method for performing regression on images. As such, its impact is dependent on the tasks for which it is used. Such tasks could have a wide range of positive benefits in numerous computer vision related fields. For example, the ability to quantify levels of disease in plants, and perhaps in other domains, may be of benefit in the agricultural and healthcare fields.
Enabling few-shot regression may further assist with 'democratizing' neural networks, in the sense that the method could enable smaller businesses or research groups in performing regression in scenarios where labeled data may be out of their means.

On the other hand, our model could be used in applications which violate privacy, for example by facilitating the collection of information such as age or sentiments from individual photographs.

Furthermore, our model is sensitive to the same biases found in the data used to train the GAN - and in the case of natural-language based regression, also to the biases present in CLIP. As such, it may assist in perpetuating biases such as gender norms (as seen in the makeup sorting experiment) or racial discrimination (\eg Asian descent is correlated with age in the data set, and this is reflected in model predictions).

\section{Licenses and privacy}
The datasets and models used in our work and their respective licenses are outlined in Table \ref{tbl:licenses}.

\begin{table}[!hbt]
    \small
    \centering
    \caption{Datasets and models used in our work and their respective licenses.}
    \makebox[\linewidth]{
    \begin{tabular}{c c}
    
    \begin{tabular}{l c c}
        Dataset & Source & License \\ \hline
        FFHQ & \cite{karras2019style} & \ccbyncsa \footnote{Individual images under different licenses. See \url{https://github.com/NVlabs/ffhq-dataset}}\\
        CelebA-HQ & \cite{karras2017progressive} & \ccbync \\ 
        AFHQ & \cite{choi2020starganv2} & \ccbync \\
        Stanford Cars & \cite{krause20133d} & \href{https://image-net.org/download.php}{ImageNet License} \\
        CompCars & \cite{yang2015large} & \href{http://mmlab.ie.cuhk.edu.hk/datasets/comp_cars/}{ Non-Commercial Research} \\ 
        CACD & \cite{chen14cross} & \href{https://bcsiriuschen.github.io/CARC/}{Academic Research} \\ 
        PlantVillage & \cite{hughes2015open} & \cczero \\
    \end{tabular}
    &
    \begin{tabular}{l c c}
        Model & Source & License \\ \hline
        StyleGAN2 & \cite{karras2020analyzing} & \nvsrc \\ 
        GHFeat & \cite{xu2020generative} & No License \\ 
        SSV & \cite{mustikovela2020self} & \nvsrc \\ 
        Scikit-Learn & \cite{pedregosa2011scikit} & \bsd \\
        WHENet & \cite{zhou2020whenet} & \bsd \\
        DEX & \cite{Rothe-ICCVW-2015} & No License \\
        pSp & \cite{richardson2020encoding} & \mitlic \\
        e4e & \cite{tov2021designing} & \mitlic \\
        ReStyle & \cite{alaluf2021restyle} & \mitlic \\
        InterFaceGAN & \cite{shen2020interpreting} & \mitlic \\
        SeFa & \cite{shen2020closedform} & \mitlic \\
        StyleCLIP & \cite{patashnik2021styleclip} & \mitlic \\
        CLIP & \cite{radford2021learning} & \mitlic \\
        PoseContrast & \cite{Xiao2020PoseContrast} & \mitlic \\
        FSA & \cite{yang2019fsa} &  \href{https://www.apache.org/licenses/LICENSE-2.0}{Apache License V2.0} \\ 
        StyleGAN2-pytorch & \cite{rosinalitySG2} & \mitlic \\
        StyleGAN-ADA & \cite{Karras2020ada} & \href{https://nvlabs.github.io/stylegan2-ada-pytorch/license.html}{Nvidia Source Code License} \\
    \end{tabular}
    \\
    (a) Datasets & (b) Models
    \end{tabular}
    }
    \label{tbl:licenses}
\end{table}

Some of the datasets in use, an in particular FFHQ \cite{karras2019style}, CelebA-HQ \cite{karras2017progressive} and CACD \cite{chen14cross}, contain personally identifiable data in the form of face images.

We have not reached out to receive consent from the individuals portrayed in the images. However, all three image sets are composed of publicly available celebrity images or faces of individuals crawled from flicker, all of which were uploaded under permissive licenses which allow free use, redistribution, and adaptation for non-commercial purposes. The curators of all sets provide contact details for individuals who wish to have their images removed from the set.

\section{Ablation}
We conduct an ablation study on several components of our method. Namely, our choice of inversion mechanism and the importance of our layer-weighting approach (as detailed in \subsecref{subsec:bridging} of the core paper).

\subsection{Inversion comparisons}
We evaluate the performance of our model when utilizing different inversion methods in order to obtain the latent code for both train and test images. We compare our results on the human face pose and age estimation tasks using the CelebA-HQ \cite{karras2017progressive} dataset. Specifically, we compare four \textit{GAN Inversion} encoder models - pSp \cite{richardson2020encoding}, e4e \cite{tov2021designing}, ReStyle-psp and ReStyle-e4e \cite{alaluf2021restyle} which uses the former encoders in an iterative refinement scheme. In all cases we use the official pre-trained models provided by the authors.

\begin{figure}
\setlength{\tabcolsep}{1pt}
    \centering
    \begin{tabular}{c c}
        \includegraphics[width=0.5\linewidth]{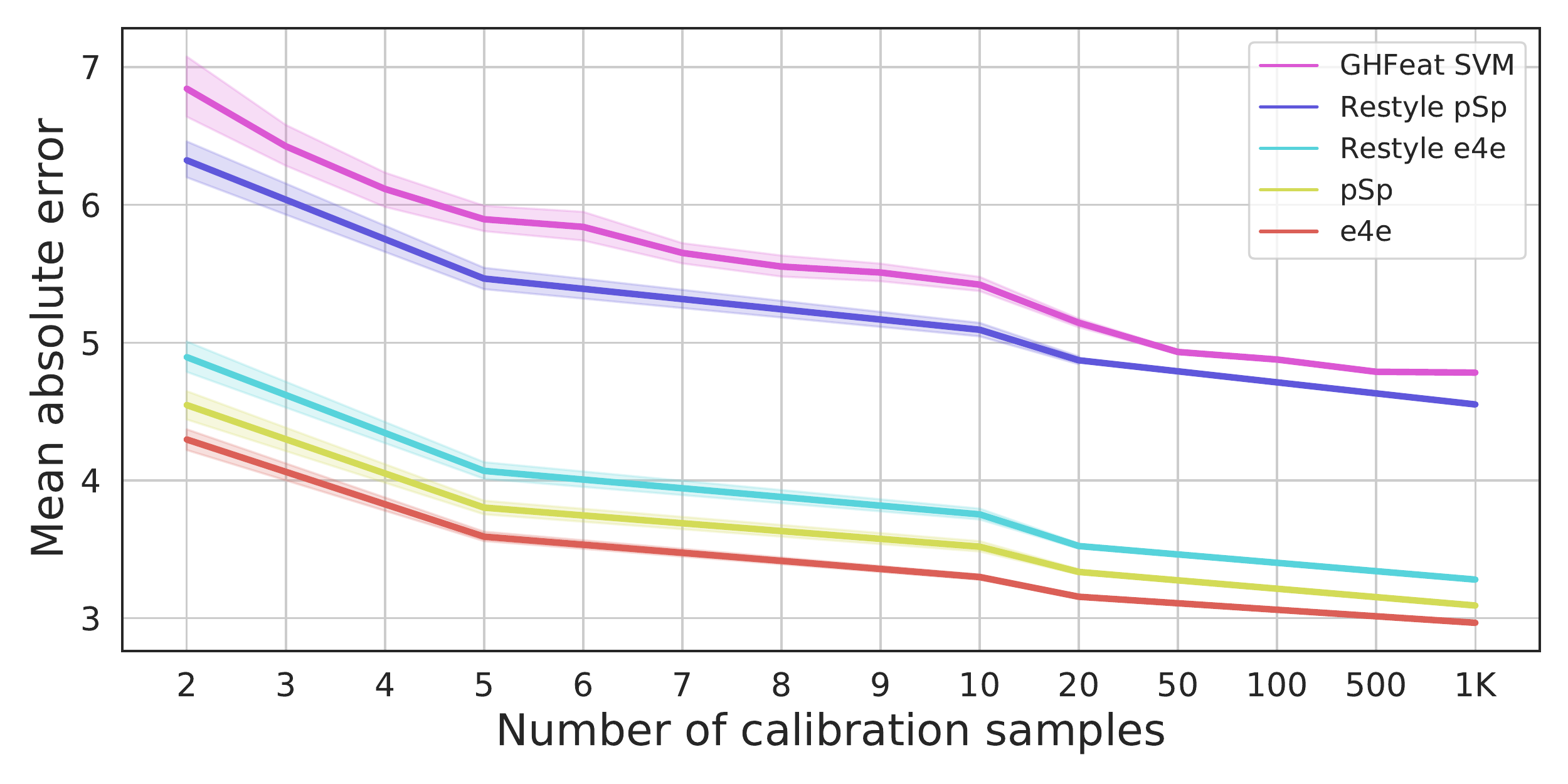} & \includegraphics[width=0.5\linewidth]{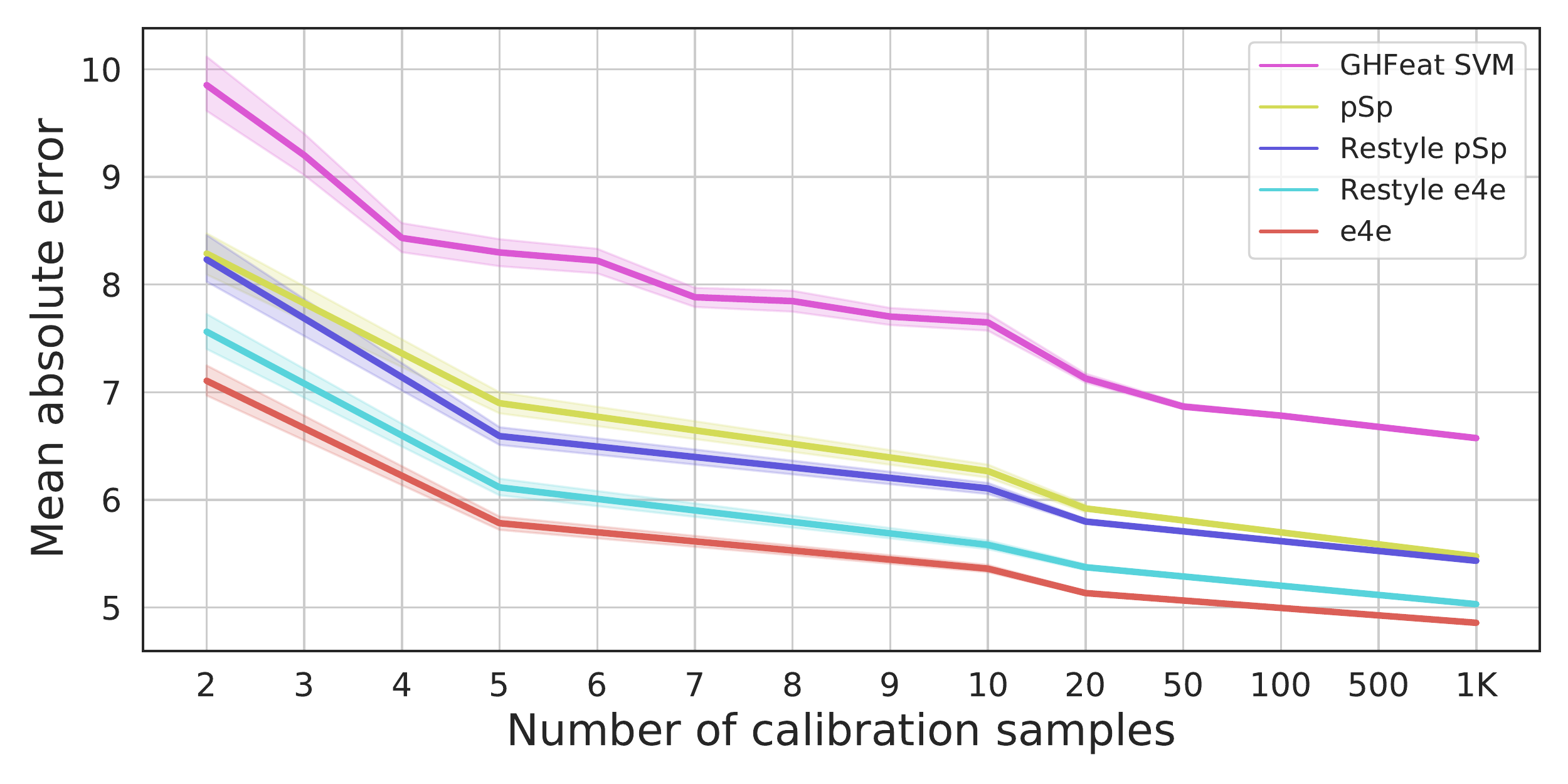} \\
        (a) Pose & (b) Age
    \end{tabular}
    \caption{
    Quantitative comparisons of four different \textit{GAN Inversion} encoders: e4e \cite{tov2021designing}, psp \cite{richardson2020encoding} and ReStyle e4e and ReStyle \cite{alaluf2021restyle} on the CelebA-HQ dataset \cite{karras2017progressive}.
    }
    \vspace{-5pt}
    \label{fig:comparing_inversion_methods}
\end{figure}

For reference, we also include the most competitive baseline, GHFeat-SVM which was introduced in \subsecref{subsec:feature_space}. As a reminder, this baseline was devised based on the same latent-distance principles but applied in the feature space of GHFeat \cite{xu2020generative}.
Results are displayed in \figref{fig:comparing_inversion_methods}.
As can be seen, our method outperforms GHFeat-SVM regardless of the choice of inversion method. 
Furthermore, e4e \cite{tov2021designing} is consistently superior to other methods, and e4e-based methods are superior to pSp-based methods.
Tov \etal \cite{tov2021designing} demonstrated that there exists a trade-off between distortion and editability. This tradeoff stems from the ability to invert images into more semantic regions of the latent space. While the e4e encoder tends towards preserving such semantics, pSp is trained with the sole purpose of image reconstruction. As our method relies heavily on such latent space semantics, it is unsurprising to see a consistent, even though minor, advantage towards e4e-based methods.

While superior results are obtained with all inversion methods, we conclude that our method works best with semantic preserving encoders and recommend using such.

\subsection{Bridging the gap}

We evaluate the contribution of our per-layer latent direction weighting approach, described in \subsecref{subsec:bridging} of the main paper. 

We compare our proposed approach with three alternatives for computing distances between latent codes in \wplus and boundaries found in \w.
First, we use a simple model dubbed ``All Layers" in which we compute the distance of the latent code in each separate layer to the same \w-space boundary. We then use all such distances ($18$ in total) as features for the regression model. Second, we consider a model dubbed ``Euclidean", in which we duplicate the boundary over all layers and compute a simple Euclidean distance.
Last, we consider a model which uses only the distance along the single layer which provides optimal performance. In the case of poses, for example, this is layer $2$. Note that determining the optimal layer in this manner requires a large continuously tagged dataset to evaluate against, which may not be feasible in practical applications.
Our own model uses a weighted distance metric where the contribution of each layer is scaled according to our semantic-mapping importance scores determined in an unsupervised manner.

\begin{figure}[h]
\centering
\includegraphics[width=0.5\textwidth]{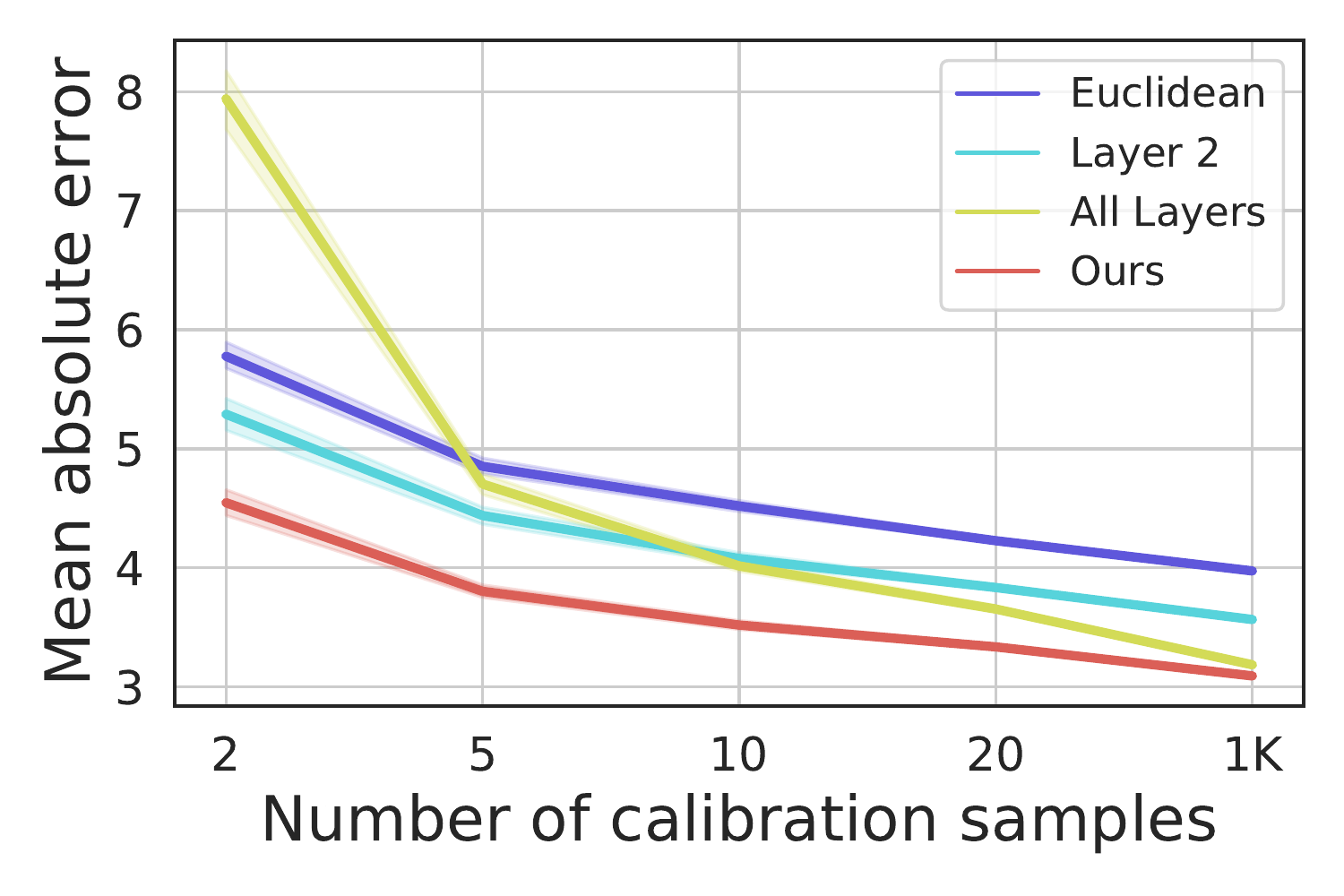}
\caption{Comparing several approaches for calculating latent-space distances between codes in \wplus and boundaries in \w. The ''Euclidean" model calculates the Euclidean distance between the latents and a boundary obtained by replicating the \w boundary along all layers of \wplus. ''All Layers" calculates a per-layer distance and uses all 18 distances as features. ``Layer 2" uses only the distance calculated on layer 2 of the latent code, which was experimentally observed to provide the best single-layer results for pose. Finally, our model uses a weighted distance as outlined in \subsecref{subsec:bridging} of the main paper. As can be seen, our proposed method is superior to other method in the few-shot domain and is matched by ``All Layers" only when provided with a thousand labeled samples.}
\label{fig:bridge_ablation}
\end{figure}

The performance of our method and all alternative is reported in \figref{fig:bridge_ablation}.
Our weighted model consistently outperforms the other alternatives in low-supervision settings, and achieves similar results to the ``All Layers" model with extensive supervision. This shows that our weighted distances accurately reflect the importance of the distance across each layer, without having to rely on the additional supervision required to determine such a weighting with multiple-feature regression.

\begin{figure}[hb!]
\centering
\includegraphics[width=0.9\textwidth]{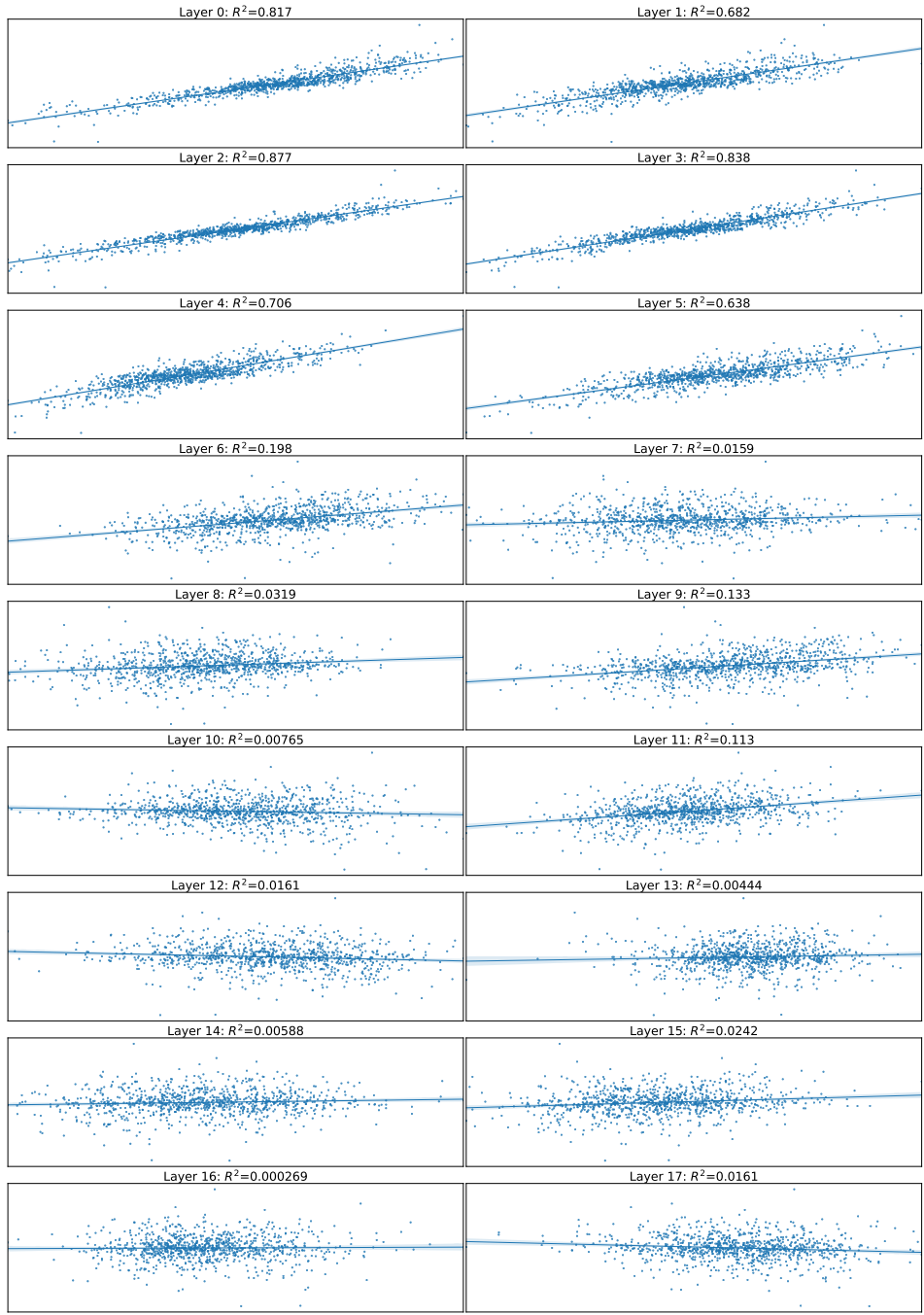}
\caption{Measuring the linear correlation of yaw angle with distance from hyperplane for each layer separately. In each subplot, the x-axis is the distance of this layer in the latent code from the boundary while the y-axis is the ground truth yaw angle. As can be seen, distance in first layers are better linearly correlated to head pose than last layers.}
\label{fig:layer_linearity}
\end{figure}

\begin{figure}
\setlength{\tabcolsep}{1pt}
    \centering
    \begin{tabular}{c c}
        \includegraphics[width=0.5\linewidth]{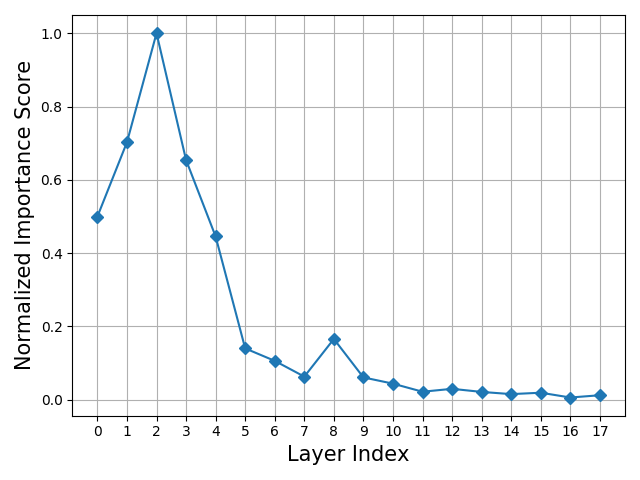} & \includegraphics[width=0.5\linewidth]{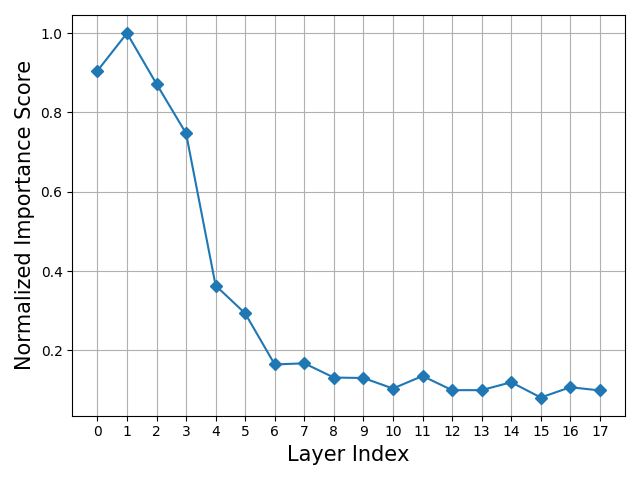} \\
        (a) Un-normalized & (b) Normalized
    \end{tabular}
    \caption{
    The results of our per-layer importance scores approach as outlined in \subsecref{subsec:bridging}, for the head pose attribute. (a) Un-normalized importance scores, before accounting for the scale of gradients in each layer. (b) Normalized importance scores, after accounting for gradient scales.
    }
    \vspace{-5pt}
    \label{fig:bridge_weights_pose}
\end{figure}

Beside obtaining superior performance, we next demonstrate that our approach in fact identifies layers that are highly correlated with a semantic attribute. For this end, we first fit a set of linear models for the pose estimation task, using the individual distances along each layer of \wplus, one at a time. The results and their $R^2$ coefficients are shown in \figref{fig:layer_linearity}. \rg{Note that obtaining accurate correlation scores in this manner requires a large labeled dataset and may hence be unfeasible.}
Next, in \figref{fig:bridge_weights_pose} we show the layer importance scores extracted by our unsupervised method. As can be seen, our unsupervised layer scoring approach successfully identifies layers with high correlation to the semantic property.  

\subsection{Comparing against generating a continuously-labeled data}
As discussed in the paper, a popular approach to using Deep Generative Models for discriminative task is to generate labeled datasets and use them for training. Although our method provides means to perform regression directly in the latent space, it may also be used to generate labeled datasets. After calibrating the model, we can apply it to any latent code, including codes simply sampled from the Gaussian prior. We can thus generate a new dataset from randomly sampled codes, and dictate their labels by the latent regression model.

We perform an experiment where we generate such a dataset for human head pose and train a regression CNN directly on the generated images. To make sure the attribute varies enough in the generated dataset we perform the following process. We sample a latent code $w \in \mathcal{W}$ and a scalar $\alpha \in [-9,9]$ which was observed to be the maximal range for which the generated image does not degrade in quality. We then edit the sampled latent code by applying $w' = w + \alpha \vec{n}$. Now, we generate the image $I' = G(w')$ and infer the head pose by inputting $\alpha$ to the regression model. For the regression model, we use a model trained with 1000 labeled samples. Repeating this process 45K times, we now posses a continuously-labeled, roughly balanced generated dataset. We train multiple CNN backbones for the task of regression and test them over the annotated CelebA-HQ test set \cite{karras2017progressive}. The lowest Mean Average Error, $3.23 \pm 3.67$, was obtained with EfficientNet-b3 \cite{tan2019efficientnet}. For comparison, applying our approach with exactly the same set and supervision obtained $2.97 \pm 2.76$. 

We conclude that it is preferable to apply our method directly to regress test images, rather than generating a labeled set for downstream training. We speculate that a possible explanation for the degradation in performance is the domain gap. In the generated-dataset approach, the classifier is trained on a generated dataset and tested on a real one, without adaptation. On the other hand, in our approach, the \textit{GAN Inversion} encoder may mitigate some of that gap. Additionally, the generator and inversion encoder are trained once per-domain while the latent and CNN regressor as well as the data generation happens once per attribute. As a result, our approach, requiring just the training of a latent simple regression model requires roughly x1000 less time to train.


\section{Additional results}
\subsection{Calibrated results - Cars}
We repeat the pose experiment of section \ref{subsec:calibrated_results} on the car image domain. We compare our model to SSV. Our model utilizes the official StyleGAN2 \cite{karras2020analyzing} LSUN Car \cite{yu2015lsun} pretrained checkpoint, and the official e4e \cite{tov2021designing} inversion encoder trained on the train split of Stanford Cars \cite{krause20133d}. The semantic boundary was extracted using InterFaceGAN \cite{shen2020interpreting}. SSV was trained as in the original paper, using the CompCars dataset \cite{yang2015large}. Both models were evaluated on the test split of Stanford Cars, with pose labels acquired through Pose Contrast \cite{Xiao2020PoseContrast}. After labeling the test-set images, we discarded all images with yaw angles exceeding $90^\circ$ in either direction, \ie we evaluated only on images for which $\theta_{yaw} \in \left[-90^\circ, 90^\circ \right]$.

\begin{figure}[h]
\centering
\includegraphics[width=0.7\textwidth]{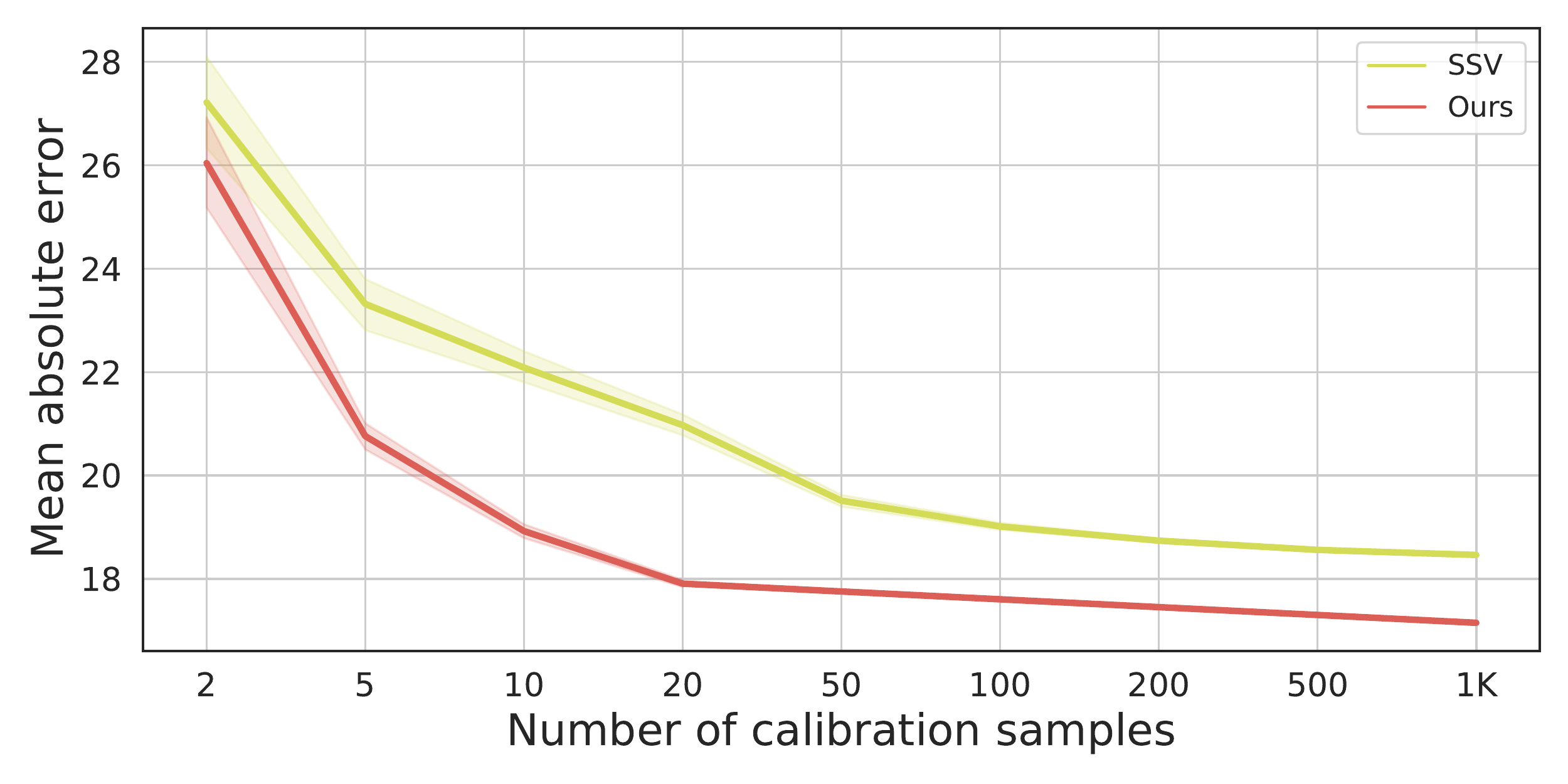}
\caption{Pose estimation error comparisons on the Stanford Car \cite{krause20133d} dataset tagged by Pose Contrast \cite{Xiao2020PoseContrast}, as a function of the number of labeled images used for calibration.}
\label{fig:compare_pose_cars}
\end{figure}

The results are shown in \figref{fig:compare_pose_cars}. Our model outperforms SSV over all tested supervision ranges, indicating that it can generalize well to the car domain.

\subsection{Uncalibrated results}

We show additional sorting results on cats and dogs, using a StyleGAN-ADA \cite{Karras2020ada} model trained on the AFHQ-dog dataset \cite{choi2020starganv2}.

The semantic direction and the method used to extract it are outlined in each figure.

\begin{figure}[!hbt]
\setlength{\tabcolsep}{1pt}
    \centering
    \begin{tabular}{c}
        \includegraphics[width=0.9\linewidth]{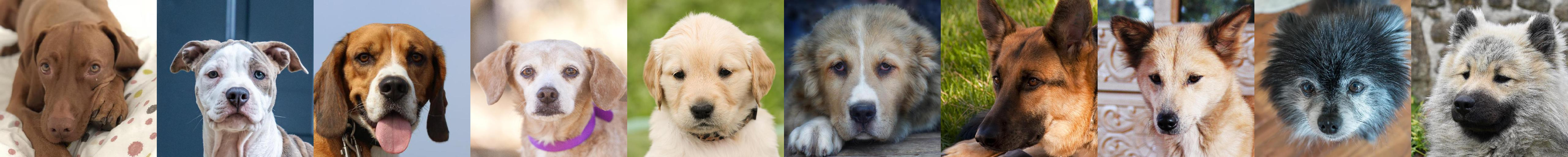} \\
        \includegraphics[width=0.9\linewidth]{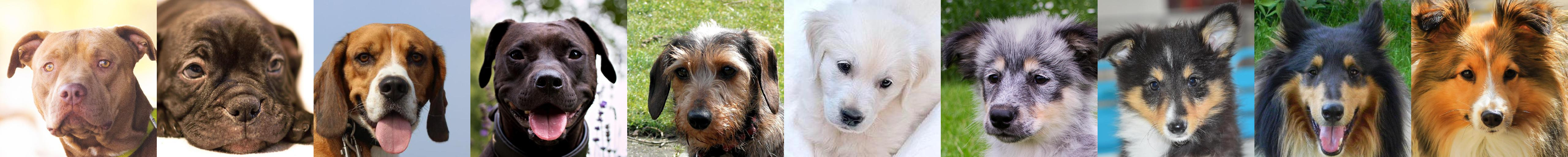} \\
        \includegraphics[width=0.9\linewidth]{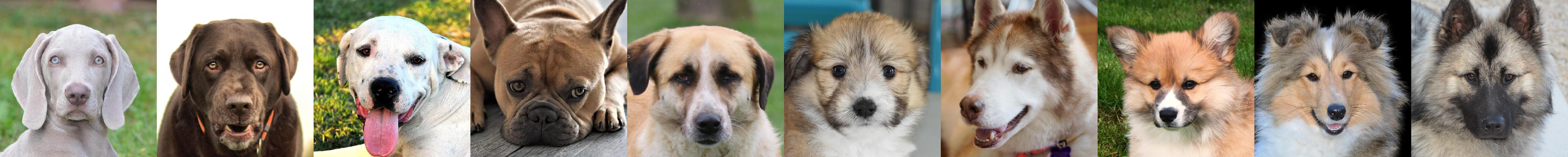} \\
    \end{tabular}
    \caption{Sorting images from AFHQ-dog \cite{choi2020starganv2} using a ``fur fluffiness" semantic directions extracted by SeFA \cite{shen2020closedform}.}
    \label{fig:dog_sort_fur}
\end{figure}

\begin{figure}[!hbt]
\setlength{\tabcolsep}{1pt}
    \centering
    \begin{tabular}{c}
        \includegraphics[width=0.9\linewidth]{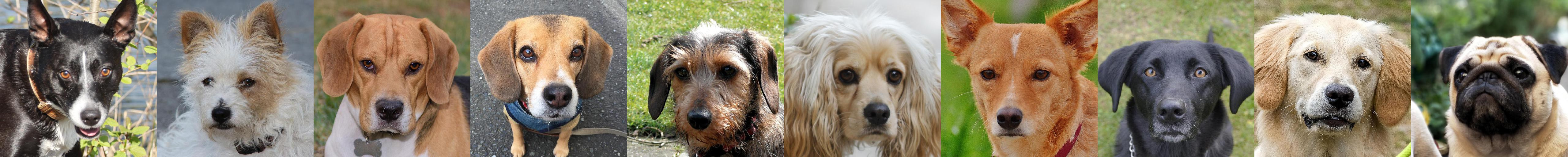} \\
        \includegraphics[width=0.9\linewidth]{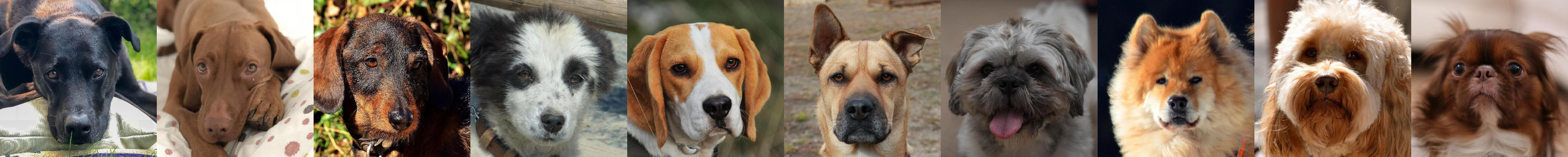} \\
        \includegraphics[width=0.9\linewidth]{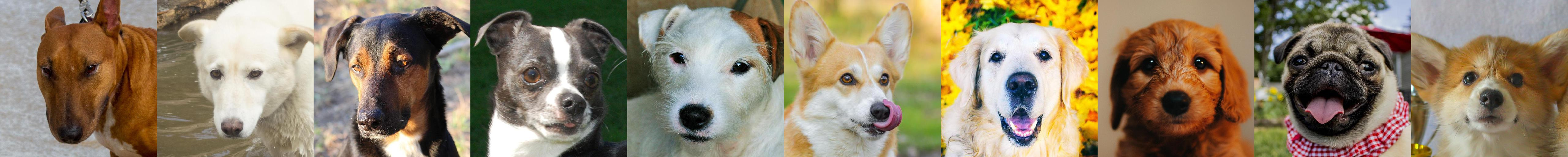} \\
    \end{tabular}
    \caption{Sorting images from AFHQ-dog \cite{choi2020starganv2} using a ``head pitch" semantic directions extracted by SeFA \cite{shen2020closedform}.}
    \label{fig:dog_sort_pitch}
\end{figure}

\begin{figure}[!hbt]
\setlength{\tabcolsep}{1pt}
    \centering
    \begin{tabular}{c}
        \includegraphics[width=0.9\linewidth]{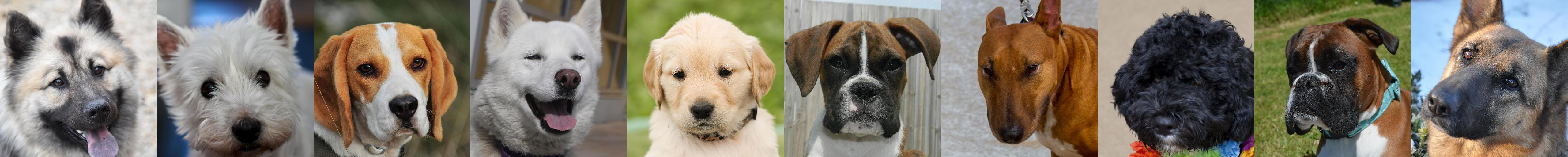} \\
        \includegraphics[width=0.9\linewidth]{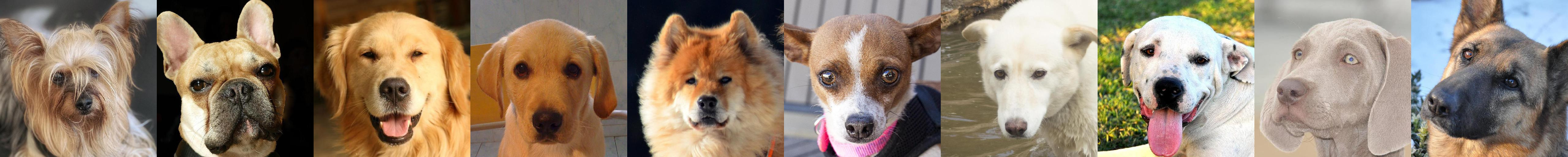} \\
        \includegraphics[width=0.9\linewidth]{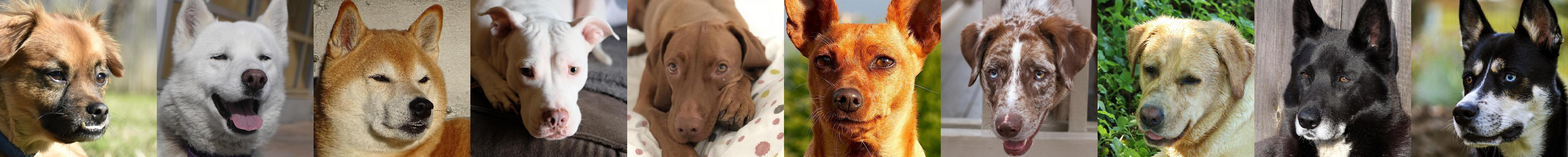}\\
    \end{tabular}
    \caption{Sorting images from AFHQ-dog \cite{choi2020starganv2} using a ``head yaw" semantic directions extracted by SeFA \cite{shen2020closedform}.}
    \label{fig:dog_sort_yaw}
\end{figure}
\begin{figure}[!hbt]
\setlength{\tabcolsep}{1pt}
    \centering
    \begin{tabular}{c}
        \includegraphics[width=0.9\linewidth]{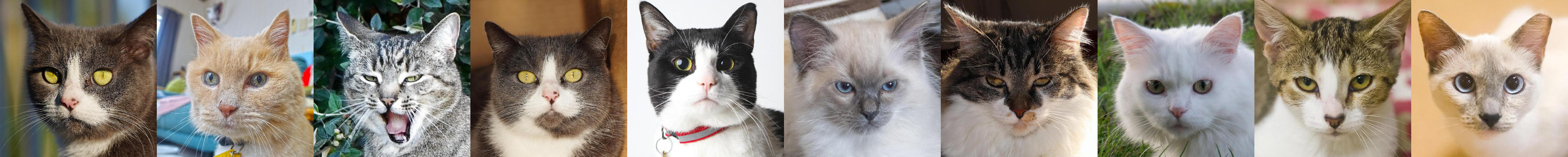} \\
        \includegraphics[width=0.9\linewidth]{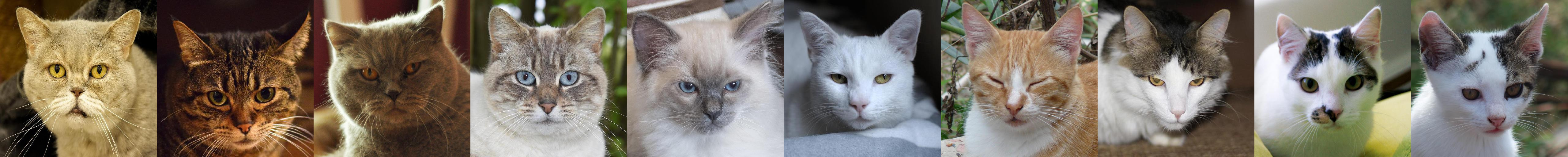} \\
        \includegraphics[width=0.9\linewidth]{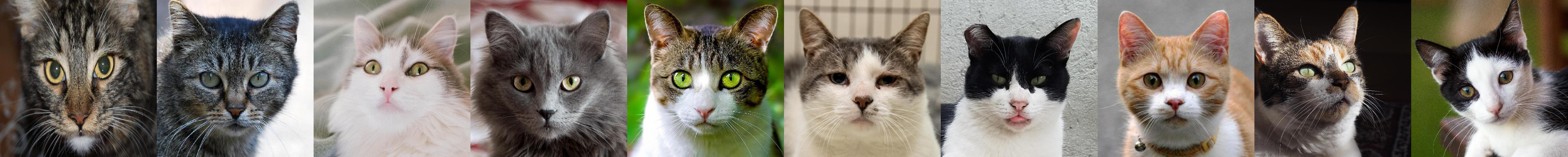} \\
    \end{tabular}
    \caption{Sorting images from AFHQ-cat \cite{choi2020starganv2} using an ``age" semantic directions extracted by SeFA \cite{shen2020closedform}.}
    \label{fig:sort_cats_age}
\end{figure}

\begin{figure}[!hbt]
\setlength{\tabcolsep}{1pt}
    \centering
    \begin{tabular}{c}
        \includegraphics[width=0.9\linewidth]{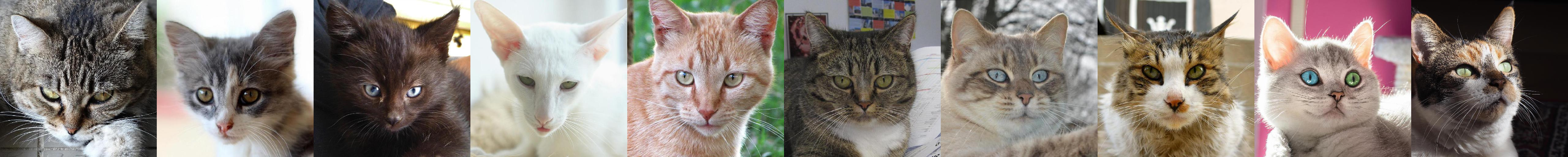} \\
        \includegraphics[width=0.9\linewidth]{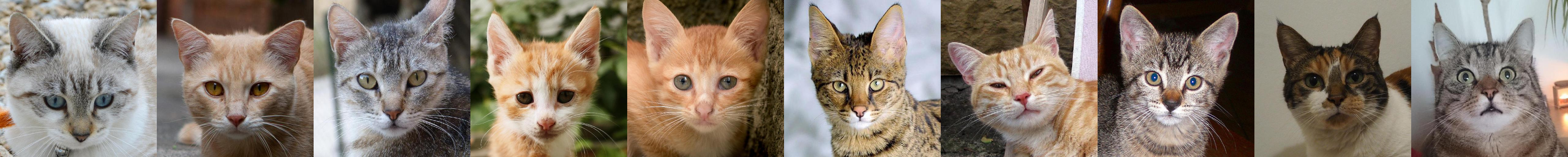} \\
        \includegraphics[width=0.9\linewidth]{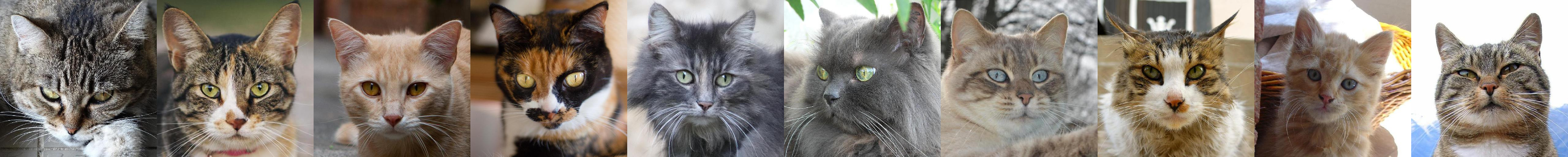}\\
    \end{tabular}
    \caption{Sorting images from AFHQ-cat \cite{choi2020starganv2} using a ``head pitch" semantic directions extracted by SeFA \cite{shen2020closedform}.}
    \label{fig:sort_10_first_afhq_cats}
\end{figure}

\begin{figure}[!hbt]
\setlength{\tabcolsep}{1pt}
    \centering
    \begin{tabular}{c}
        \includegraphics[width=0.9\linewidth]{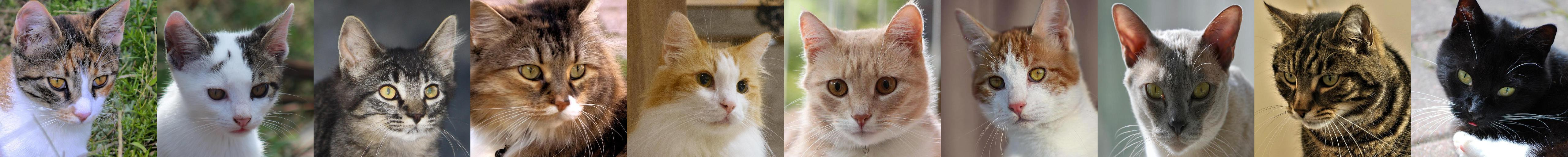} \\
        \includegraphics[width=0.9\linewidth]{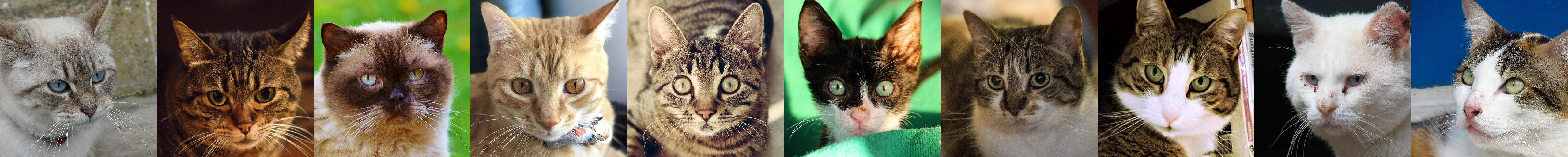} \\
        \includegraphics[width=0.9\linewidth]{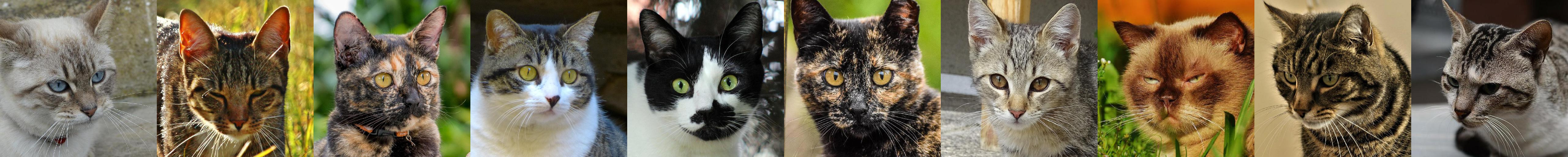} \\
    \end{tabular}
    \caption{Sorting images from AFHQ-cat \cite{choi2020starganv2} using a ``head yaw" semantic directions extracted by SeFA \cite{shen2020closedform}.}
    \label{fig:sort_10_first_afhq_cats}
\end{figure}

\begin{figure}[!hbt]
\setlength{\tabcolsep}{1pt}
    \centering
    \begin{tabular}{c}
        \includegraphics[width=0.9\linewidth]{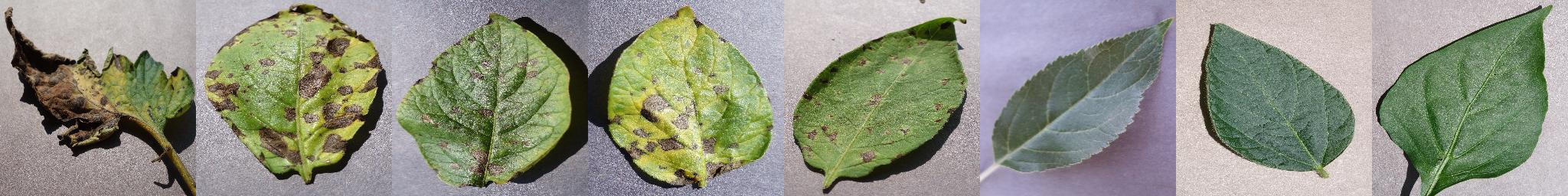} \\
        \includegraphics[width=0.9\linewidth]{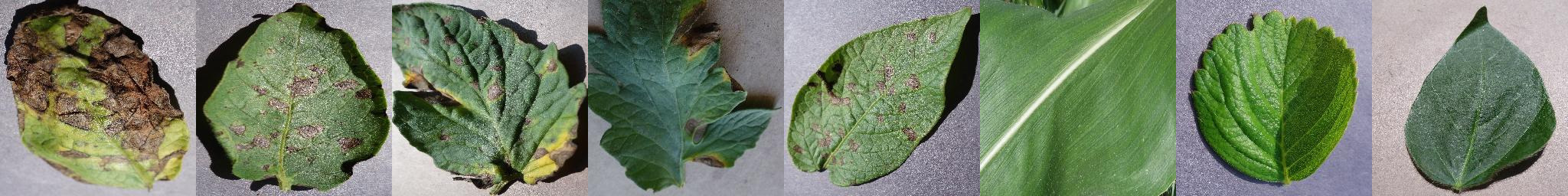} \\
        \includegraphics[width=0.9\linewidth]{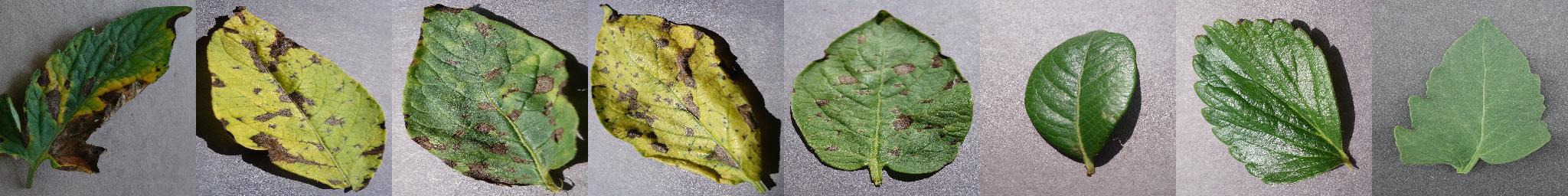} \\
    \end{tabular}
    \caption{Sorting images from Plant-Village \cite{hughes2015open} using sick-to-healthy semantic directions extracted by InterFaceGAN \cite{shen2020interpreting}. To facilitate easy visual comparisons, all sick leaves have the same disease - ``Early Blight". }
    \label{fig:sort_leaves_early_blight}
\end{figure}

\begin{figure}[!hbt]
\setlength{\tabcolsep}{1pt}
    \centering
     \begin{tabular}{c}
        \includegraphics[width=0.9\linewidth]{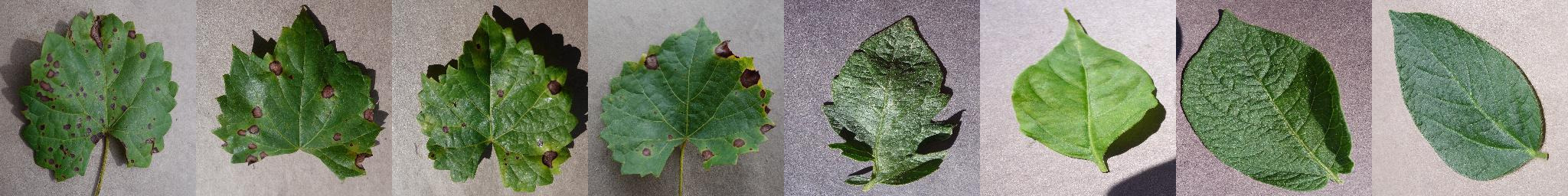} \\
        \includegraphics[width=0.9\linewidth]{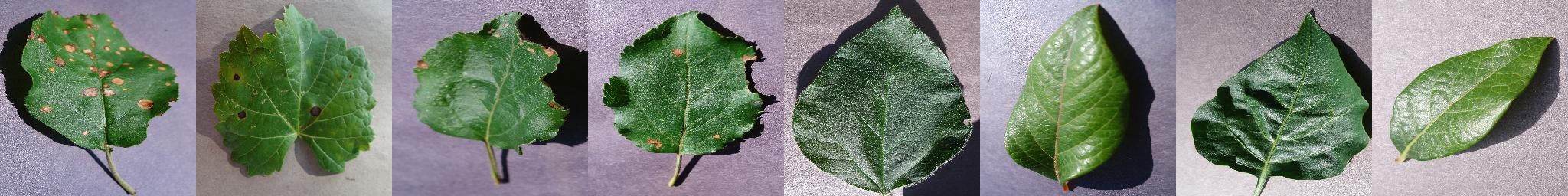} \\
        \includegraphics[width=0.9\linewidth]{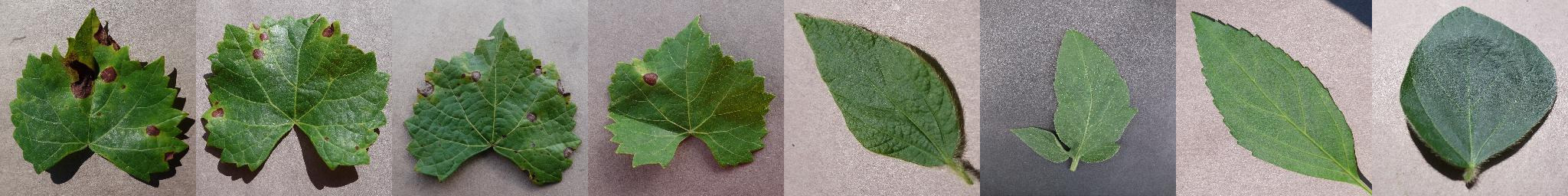} \\
    \end{tabular}
    \caption{Sorting images from Plant-Village \cite{hughes2015open} using sick-to-healthy semantic directions extracted by InterFaceGAN \cite{shen2020interpreting}. To facilitate easy visual comparisons, all sick leaves have the same disease - ``Black Rot". }
    \label{fig:sort_leaves_rot}
\end{figure}

\begin{figure}[!hbt]
\setlength{\tabcolsep}{1pt}
    \centering
     \begin{tabular}{c}
        \includegraphics[width=0.9\linewidth]{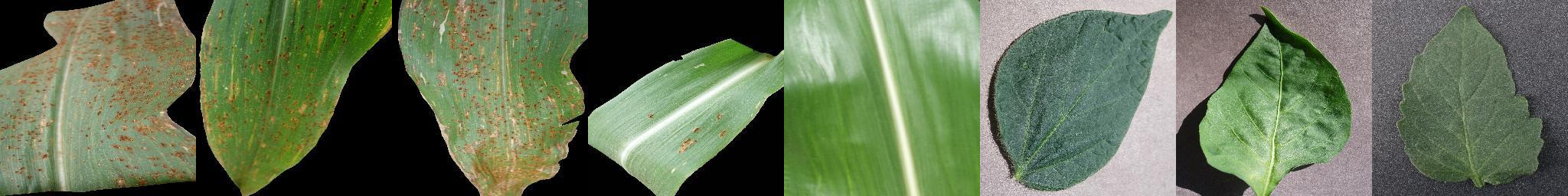} \\
        \includegraphics[width=0.9\linewidth]{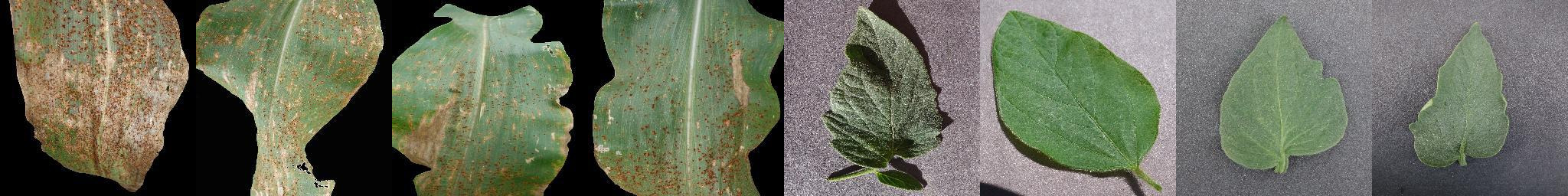} \\
        \includegraphics[width=0.9\linewidth]{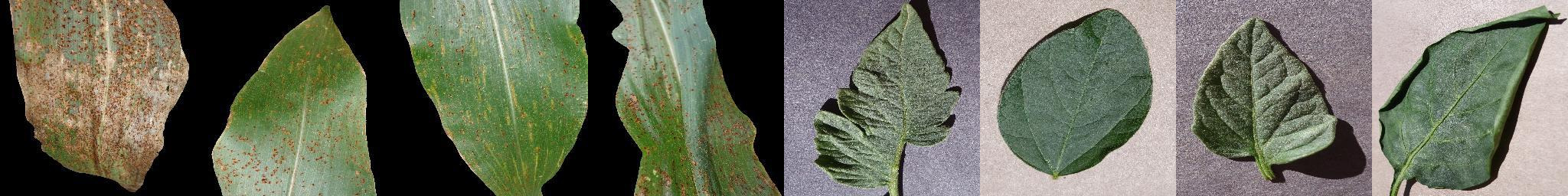} \\
    \end{tabular}
    \caption{Sorting images from Plant-Village \cite{hughes2015open} using sick-to-healthy semantic directions extracted by InterFaceGAN \cite{shen2020interpreting}. To facilitate easy visual comparisons, all sick leaves have the same disease - ``Rust". }
    \label{fig:sort_leaves_rust}
\end{figure}

\begin{figure}[!hbt]
\setlength{\tabcolsep}{1pt}
    \centering
     \begin{tabular}{c}
        \includegraphics[width=0.9\linewidth]{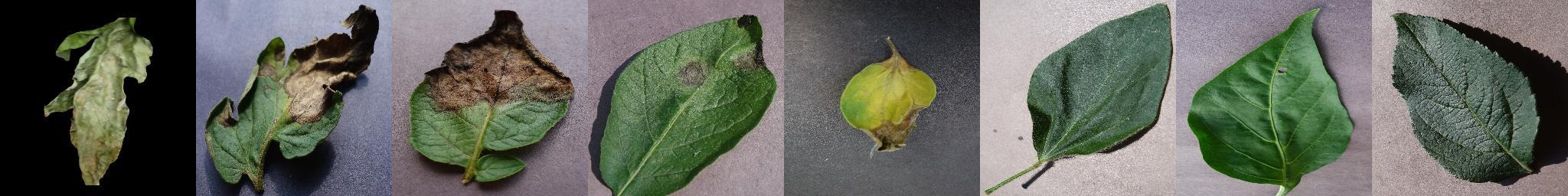} \\
        \includegraphics[width=0.9\linewidth]{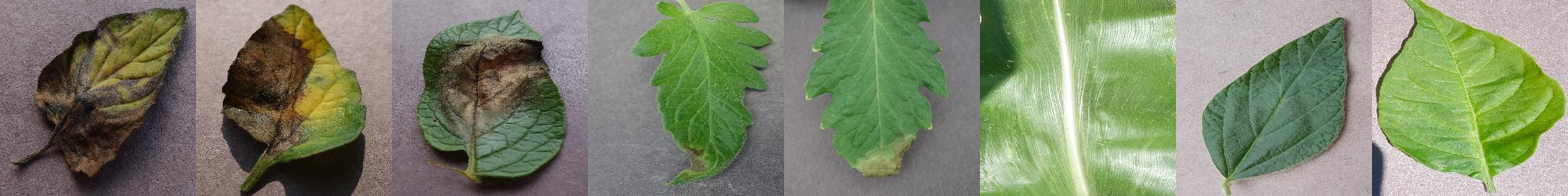} \\
        \includegraphics[width=0.9\linewidth]{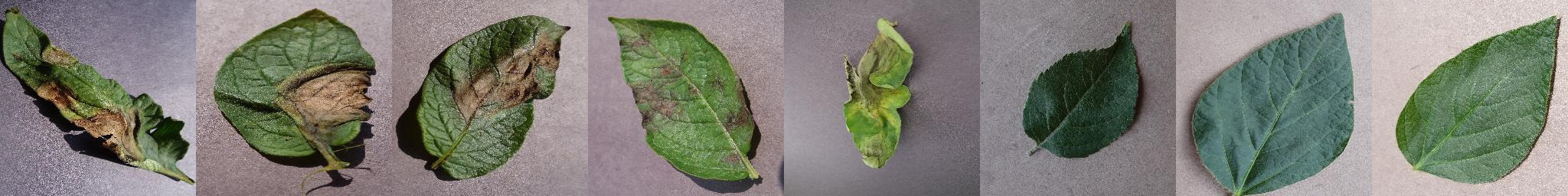} \\
    \end{tabular}
    \caption{Sorting images from Plant-Village \cite{hughes2015open} using sick-to-healthy semantic directions extracted by InterFaceGAN \cite{shen2020interpreting}. To facilitate easy visual comparisons, all sick leaves have the same disease - ``Late Blight". }
    \label{fig:sort_leaves_late_blight}
\end{figure}

\clearpage

\section{Complementing Experiments' Details}
We provide additional details about experiments conducted in the main paper.

\subsection{Accuracy of trained SVMs}
In \subsecref{subsec:feature_space} of the paper, we demonstrate that distances in the latent space of the GAN are more semantically meaningful and better behaved than equivalent distances in alternative feature spaces. For this end, we train SVMs in all feature spaces. In \tblref{tbl:baseline_svm_acc} we report the accuracy of those SVMs on validation sets.
As can be seen, the gap in performance for the task of regression cannot be easily explained by the performance of the SVM as a binary classifier. This is further evidence that StyleGAN's latent space possesses unique properties which make it suitable for the task of regression.

\begin{table}[!hbt]
    \centering
    \caption{Validation accuracy of SVM baseline models operating on pose and age, using the different feature spaces described in \subsecref{subsec:feature_space}. As can be seen, most model are decent classifiers.}
    \begin{tabular}{l|c|c}
        Feature space & Pose & Age \\ \hline
        Ours & $\mathbf{0.93}$ & $0.82$ \\ 
        Deep SVM & $0.91$ & $\mathbf{0.88}$ \\ 
        GHFeat SVM \cite{xu2020generative} & $0.91$ & $0.83$ \\ 
        PCA SVM & $0.82$ & $0.78$ \\ 
        Pixel SVM & $0.87$ & $0.74$ 
    \end{tabular}
    \label{tbl:baseline_svm_acc}
\end{table}

\subsection{What points to label?}
In order to calibrate the latent distances to actual real-world values, a few labeled samples are required. These labeled samples are then used to train a simple linear regression model. In some real-world scenarios, one won't have pre-defined disjoint sets of labeled and unlabeled samples. Rather, as most datasets form, at first the dataset is simply a collection of unlabeled samples and only later will those be annotated. While our method performs better as it gets more labeled samples, working with a few labeled samples is usually preferred. We thus provide some simple practical suggestions as to what samples one should label. Following these suggestions is increasingly more important as less points are sampled.

First we suggest to invert the unlabeled dataset to the latent space and obtain the distances from the hyperplane corresponding to the semantic latent direction, which results in a distribution of distances. This process does not require any labels. Now, we suggest choosing and labeling a set which is roughly evenly-spaced throughout the center of the distribution. There are two motivations for this sampling strategy, stemming from a single simple principle - sampling points that best represent the distribution.
First, samples on the edge of the distribution are more likely to be outliers. Latent outliers may come about when the original image is in itself an outlier in the image distribution.
Thus, discarding the noisy edges and sampling from the center of distribution is likely to better represent the dataset. Second, sampling points which are "close" to each other on the distance axis, is prone to error.
The linear relationship between the attribute and distance is modeled by $y = a \cdot d + b + \varepsilon $ where $a,b$ are the function coefficients and $\varepsilon$ is an error term.
Consider the case of sampling two points with distances $d_1, d_2$. When $\Delta d = d_1-d_2$ is small, it may be the case that $ \varepsilon > a \cdot \Delta d + b$. In such case, the noise in the observed attribute may overwhelm any signal due to the modified latent-distance, and a linear model fit to these points will fail to predict the underlying $a,b$. Sampling a set which is roughly evenly-spaced maximizes the minimal distance between any two points.

We follow these guidelines when conducting all experiments described in the paper. 
For the center, we consider $95\%$ of the data. For evenly-spaced distances we first observe that for $n$ points sampled from a uniform distribution over $[a,b]$, the minimal distance between a pair of samples is smaller or equal to $\frac{b-a}{n}$. However, choosing such a minimal distance will only allow for, at most, one sampled set. To allow greater flexibility in the choice of samples, we loosen the restriction, and sample points whose minimal allowed difference is $\frac{b-a}{n^{1.3}}$.

\subsection{Regression model regularization}
\begin{figure}[h]
\centering
\includegraphics[width=0.5\textwidth]{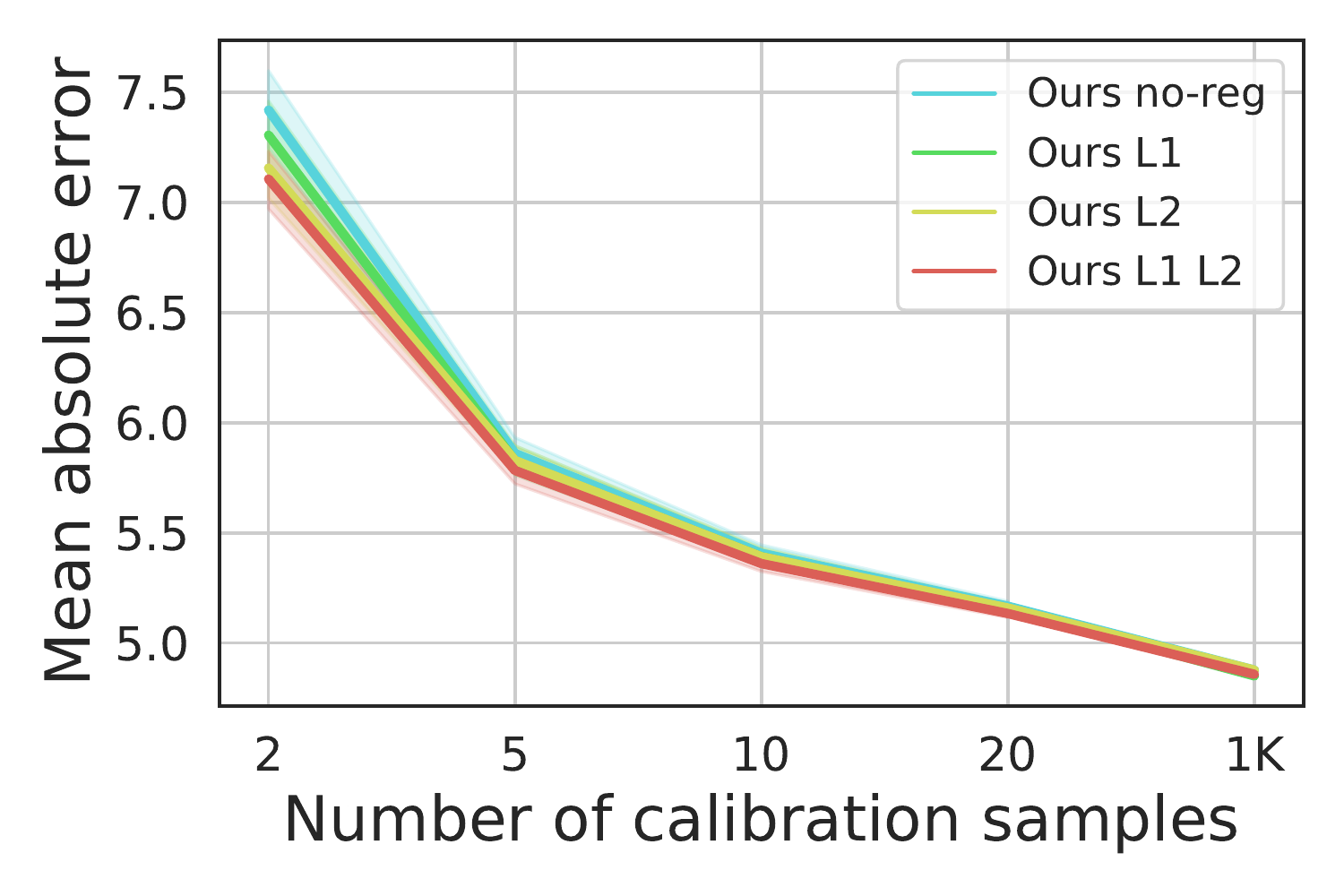}
\caption{We compare the results of our approach using different types of regularization on the simple linear regression model. As can be observed, only slight difference exist when two calibration points as used, and difference diminishes as more points are sampled.}
\label{fig:regularization_effect}
\end{figure}

Our final regression model is a simple linear regression model. However, there is still room to choose regularization. We experiment with our model without regularization, with $L1$ regularization (\ie Lasso), $L2$ regularization (\ie Ridge) or both (\ie ElasticNet). We find that there's only a slight difference in the few-shot setting and it diminishes as the number of samples increases, as demonstrated in \figref{fig:regularization_effect}. We used ElasticNet regularization in all experiments presented in the paper. We use the default penalty weighting provided by scikit-learn \cite{pedregosa2011scikit}.

\section{User Study}

In this section we provide all the details of our user study.

The user study was conducted through \href{https://freeonlinesurveys.com/}{https://freeonlinesurveys.com/}. The study was performed over a period of 5 days, with a total of 62 different responders. Individual questions saw anywhere from 20 to 62 responses (see Table \ref{tbl:study_answer_key} for exact numbers). All responders were unpaid volunteers which responded (anonymously) to a link shared among colleagues and acquaintances of the authors.

We used a three-alternative forced choice setting. Users were provided with randomly sampled sets of 10 images, sorted in three manners - once using our method, once by randomly assigning an order and once by using a dedicated baseline. For each question, the visual order of the three sorting options was randomized. Users were asked to choose the order that better matches a textual description.

For the human face domain, we used the same textual prompts as \figref{fig:clip_sort} in the main paper, with 5 randomly sampled image sets for each prompt. For a baseline, we sorted the images according to their cosine-distances from the same textual prompts directly in the CLIP \cite{radford2021learning} embedding space. For the cat domain we used the pitch and yaw directions displayed in \figref{fig:sort_afhq_cats} in the main paper and for a baseline we used pose predictions from SSV \cite{mustikovela2020self} trained on AFHQ-Cat \cite{choi2020starganv2}.

The questions and their associated image sets are shown in \figref{fig:survey_questions}. In Table \ref{tbl:study_answer_key} we provide the list of the methods used to generate each row of each question (\ie the answer key), along with the number of responders who chose each answer.

\begin{figure}[!hbt]
\centering
\makebox[\linewidth]{
\includegraphics[page=1,width=1.1\linewidth]{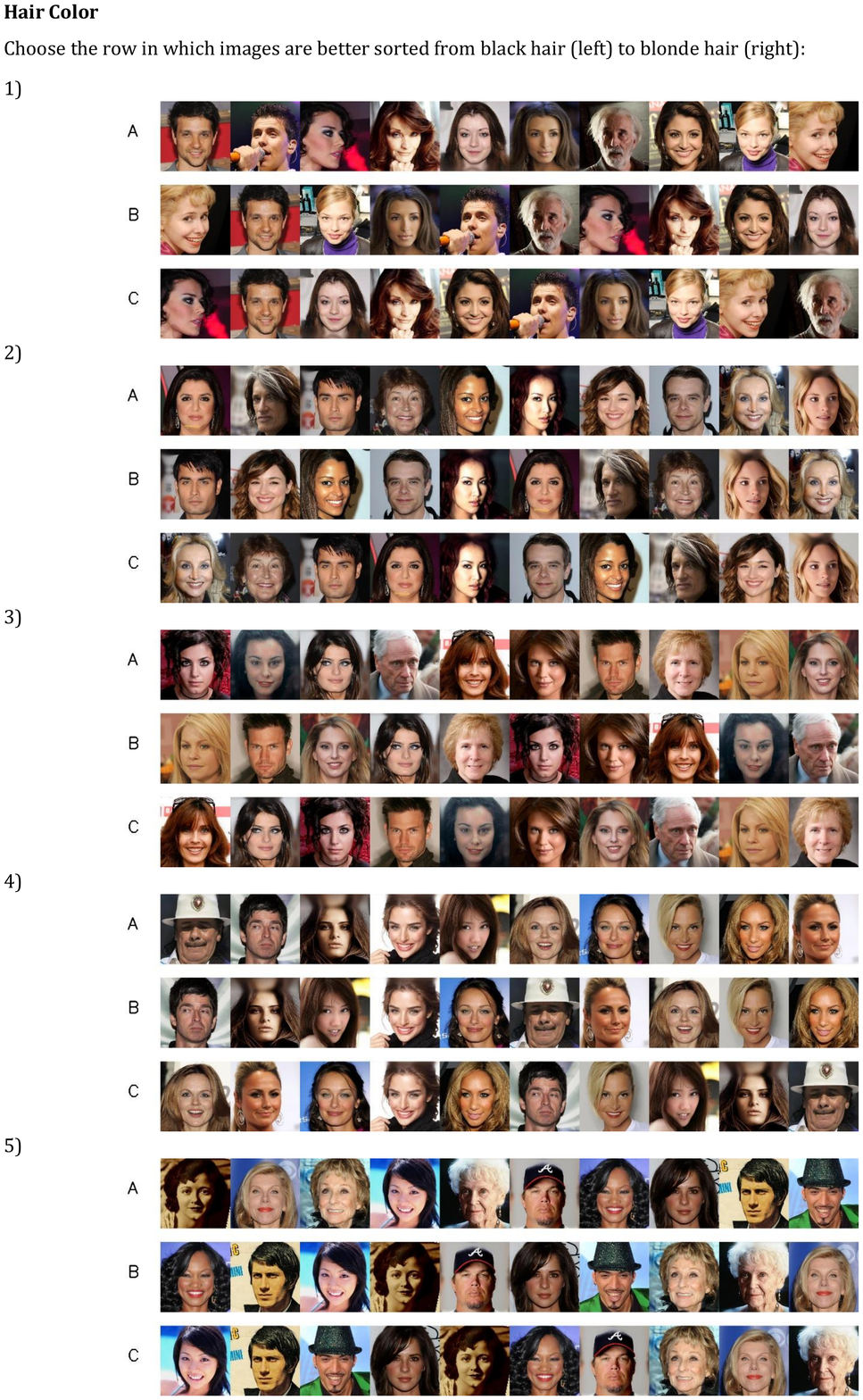}
}
\caption{All questions asked in our survey and their associated images. Page 1/6.}
\label{fig:survey_questions}
\end{figure}

\begin{figure}[!hbt]
\centering
\ContinuedFloat
\makebox[\linewidth]{
\includegraphics[page=2,width=1.1\linewidth]{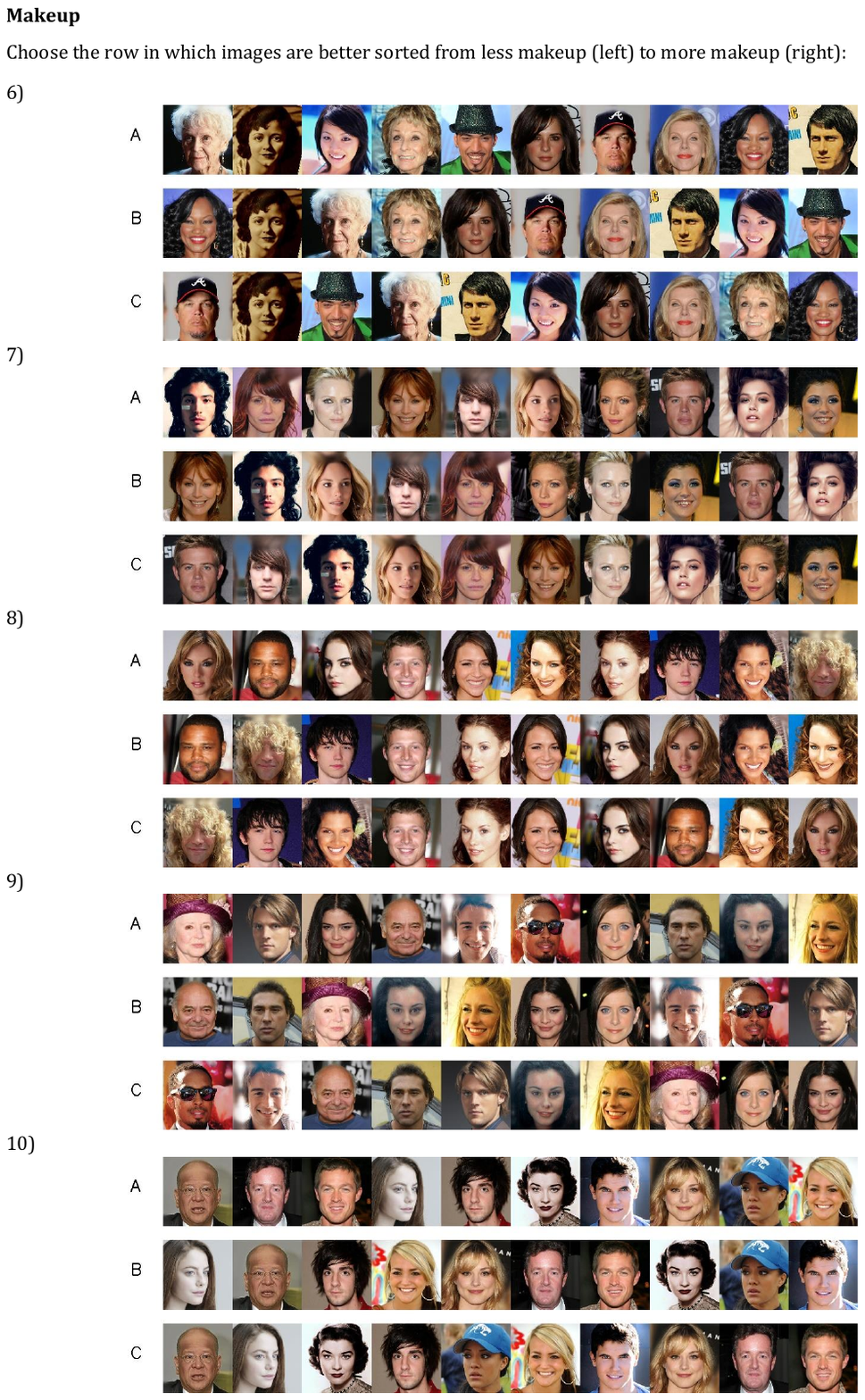}
}
\caption{All questions asked in our survey and their associated images. Page 2/6.}
\end{figure}

\begin{figure}[!hbt]
\centering
\ContinuedFloat
\makebox[\linewidth]{
\includegraphics[page=3,width=1.1\linewidth]{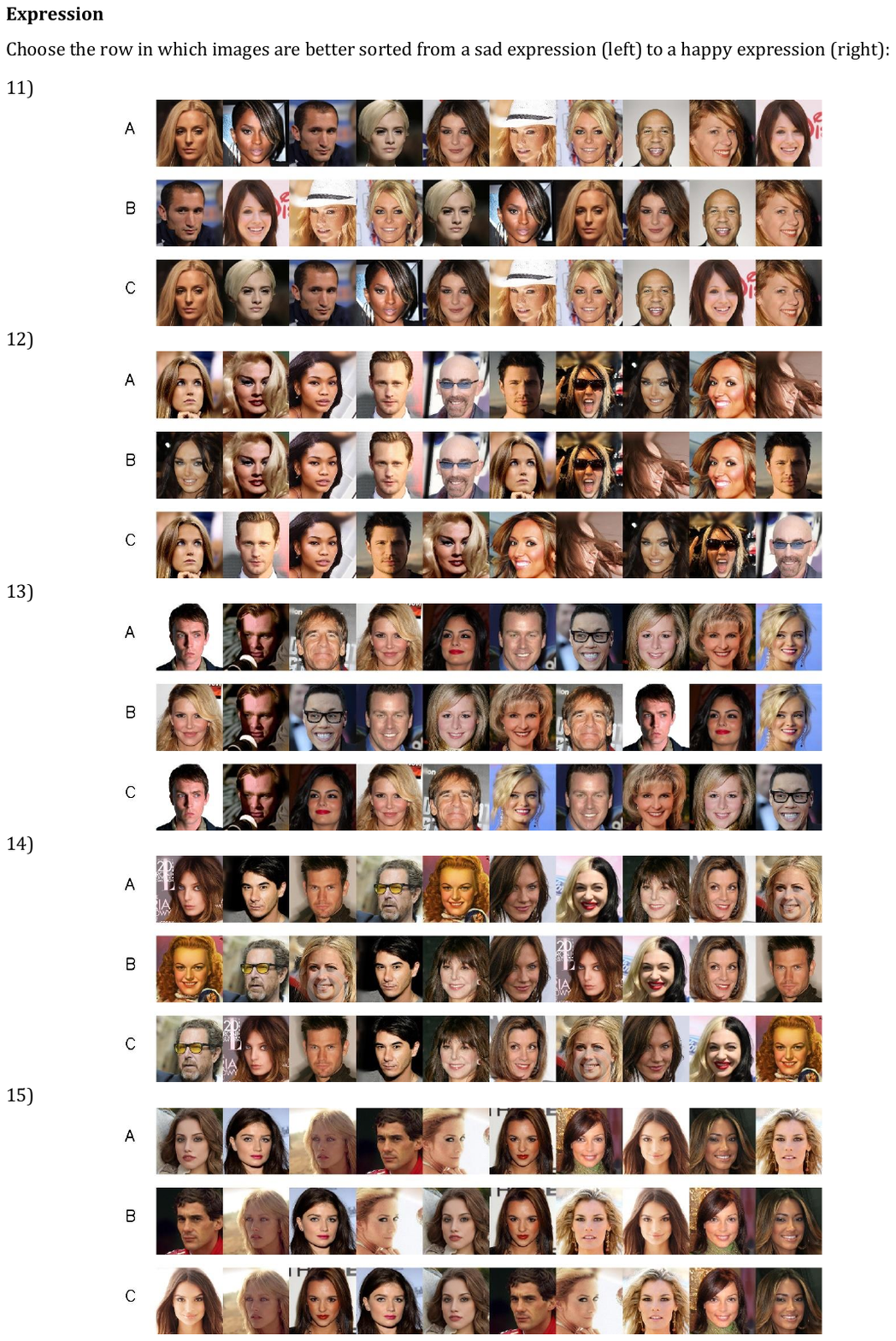}
}
\caption{All questions asked in our survey and their associated images. Page 3/6.}
\end{figure}

\begin{figure}[!hbt]
\centering
\ContinuedFloat
\makebox[\linewidth]{
\includegraphics[page=4,width=1.1\linewidth]{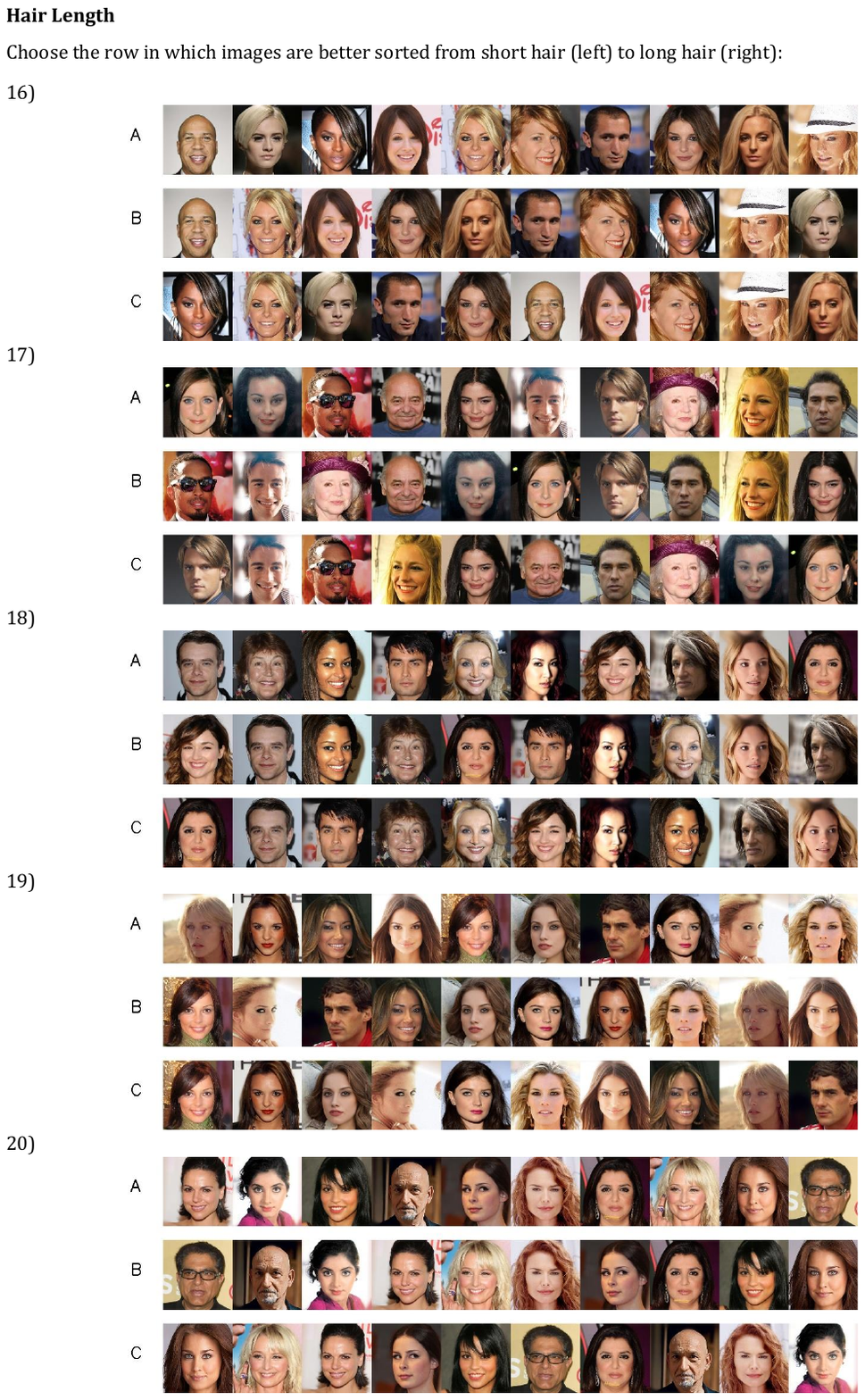}
}
\caption{All questions asked in our survey and their associated images. Page 4/6.}
\end{figure}

\begin{figure}[!hbt]
\centering
\ContinuedFloat
\makebox[\linewidth]{
\includegraphics[page=5,width=1.1\linewidth]{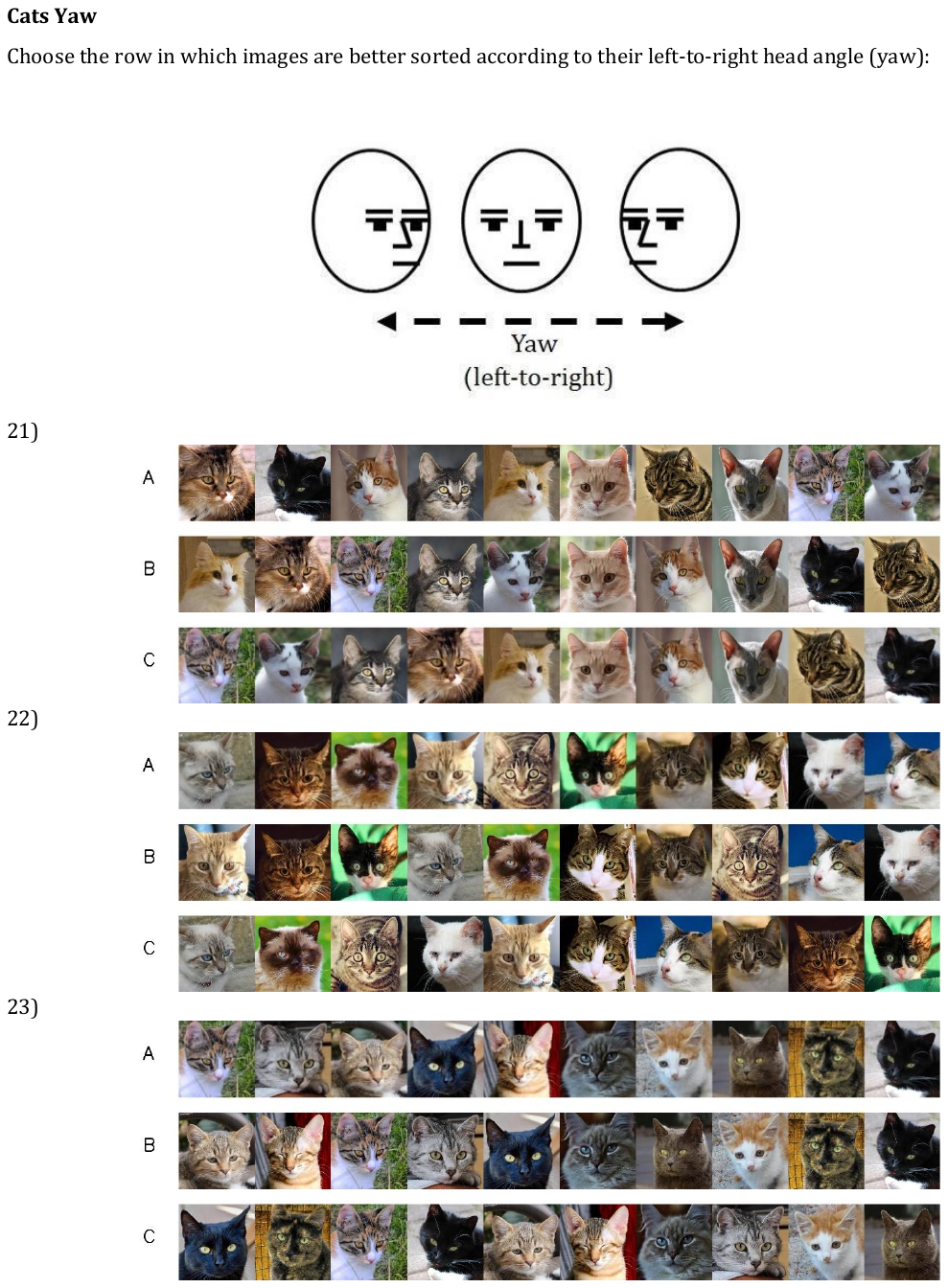}
}
\caption{All questions asked in our survey and their associated images. Page 5/6.}
\end{figure}

\begin{figure}[!hbt]
\centering
\ContinuedFloat
\makebox[\linewidth]{
\includegraphics[page=6,width=1.1\linewidth]{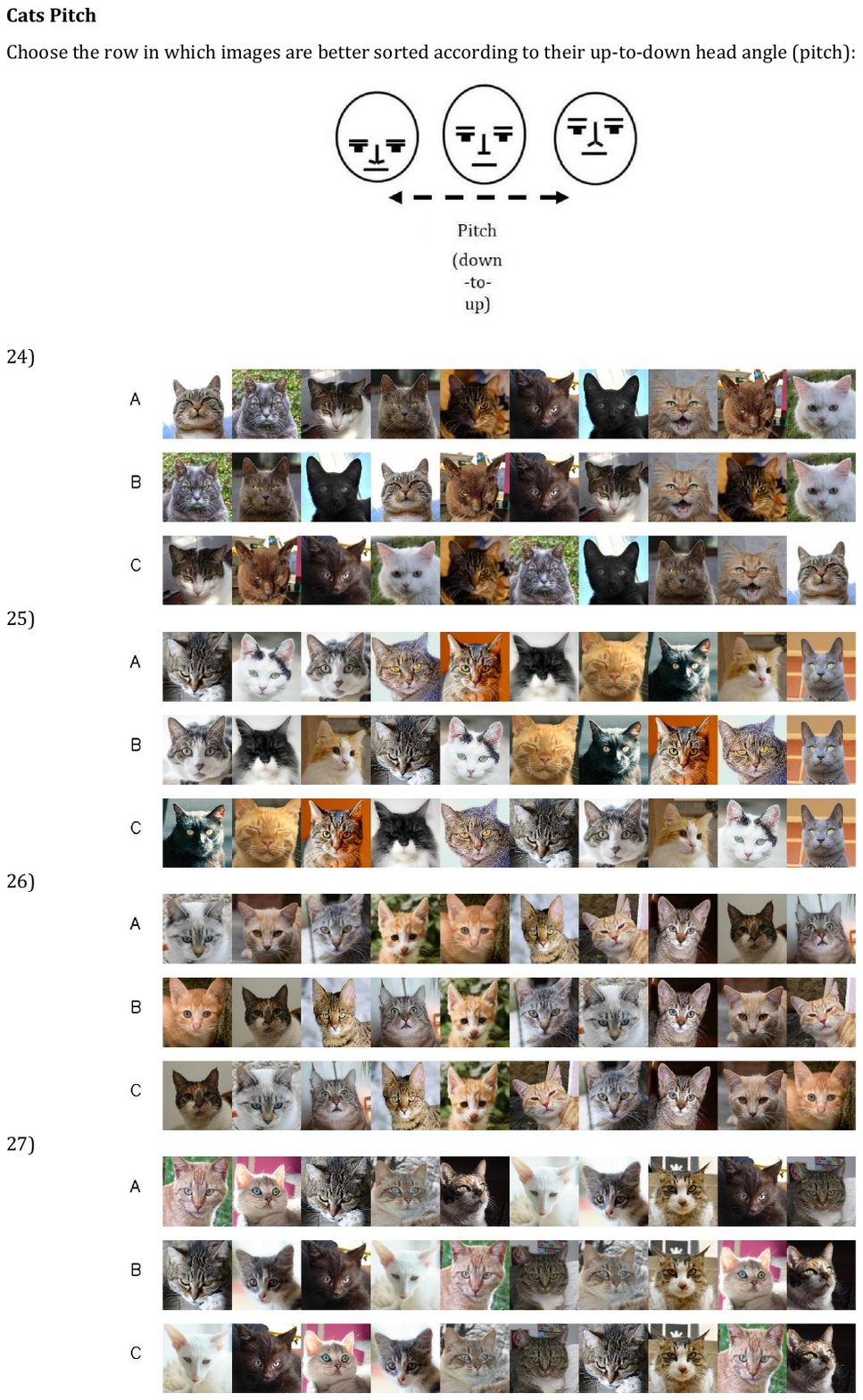}
}

\caption{All questions asked in our survey and their associated images. Page 6/6.}
\end{figure}

\begin{table}[!hbt]
    \small
    \centering
    \caption{User study answer key and the number of responders that picked each answer}
    \makebox[\linewidth]{
    \begin{tabular}{c c}

    \begin{tabular}{c c c c}
        Question & Ours & CLIP & Random \\ \hline
        \multicolumn{4}{c}{Hair color}  \\  \cdashline{1-4} \\
        1 & \bf Bottom (46) & Top (16)    & Middle (0) \\ 
        2 & \bf Middle (50) & Top (11)    & Bottom (1) \\ 
        3 & \bf Bottom (34) & Top (27)    & Middle (1) \\ 
        4 & \bf Middle (53) & Top (9)     & Bottom (0) \\ 
        5 & \bf Bottom (45) & Middle (17) & Top (0)    \\ 
        \hline
        \multicolumn{4}{c}{Makeup}  \\ \cdashline{1-4} \\
        6  & \bf Bottom (45) & Top (8)     & Middle (4)  \\ 
        7  & \bf Bottom (35) & Top (9)     & Middle (13) \\ 
        8  & \bf Middle (36) & Bottom (17) & Top (4)     \\ 
        9  & \bf Bottom (41) & Middle (11) & Top (5)     \\ 
        10 & \bf Top (44)    & Middle (7)  & Bottom (6)  \\ 
        \hline
        \multicolumn{4}{c}{Expression}  \\ \cdashline{1-4} \\
        11 & \bf Bottom (40) & Top (13)    & Middle (2) \\ 
        12 & \bf Bottom (46) & Top (9)     & Middle (0) \\ 
        13 & \bf Bottom (38) & Top (15)    & Middle (2) \\ 
        14 & Top (19)    & \bf Bottom (35) & Middle (1) \\ 
        15 & Bottom (4)  & \bf Middle (47) & Top (4)    \\ 
        \hline
        \multicolumn{4}{c}{Hair length} \\ \cdashline{1-4} \\
        16 & \bf Top (52)    & Bottom (1) & Middle (2)  \\ 
        17 & \bf Middle (46) & Top (1)    & Bottom (8)  \\ 
        18 & \bf Top (33)    & Middle (0) & Bottom (22) \\ 
        19 & \bf Middle (50) & Bottom (1) & Top (4)     \\ 
        20 & \bf Middle (52) & Top (3)    & Bottom (0)  \\ 
    \end{tabular}
    &
    \begin{tabular}{c c c c}
        Question & Ours & SSV & Random \\ \hline
        \multicolumn{4}{c}{Yaw} \\ \cdashline{1-4} \\
        21 & Bottom (4) & \bf Middle (18) & Top (1)    \\ 
        22 & \bf Top (22)   & Middle (0)  & Bottom (1) \\ 
        23 & \bf Top (20)   & Middle (2)  & Bottom (1) \\ 
        \hline
        \multicolumn{4}{c}{Pitch} \\ \cdashline{1-4} \\
        24 & \bf Bottom (17) & Middle (0) & Top (3)    \\
        25 & \bf Top (16)    & Bottom (3) & Middle (1) \\ 
        26 & \bf Top (17)    & Bottom (1) & Middle (2) \\ 
        27 & \bf Middle (16) & Bottom (2) & Top (2)    \\ 
    \end{tabular}
    \\
    (a) Human faces & (b) Cats
    \end{tabular}
    }
    \label{tbl:study_answer_key}
\end{table}


\end{document}